\documentclass{article}

\usepackage[preprint,nonatbib]{neurips_data_2024}
\usepackage[numbers]{natbib}

\usepackage[utf8]{inputenc} 
\usepackage[T1]{fontenc}    
\usepackage{hyperref}       
\usepackage{url}            
\usepackage{booktabs}       
\usepackage{amsfonts}       
\usepackage{nicefrac}       
\usepackage{microtype}      
\usepackage{xcolor}         
\usepackage{wrapfig}

\usepackage{algorithm}
\usepackage{algpseudocode}

\usepackage{caption}
\usepackage{subcaption}

\usepackage{amsmath,amsfonts,bm}

\usepackage{amsmath}
\usepackage{amssymb}
\usepackage{mathtools}
\usepackage{amsthm}

\usepackage{todonotes}

\theoremstyle{plain}
\newtheorem{theorem}{Theorem}[section]

\theoremstyle{definition}
\newtheorem{definition}[theorem]{Definition}

\theoremstyle{remark}

\newtheoremstyle{named}{}{}{\itshape}{}{\bfseries}{.}{.5em}{\thmnote{#3}}
\theoremstyle{named}
\newtheorem*{namedtheorem}{Theorem}

\newcommand{\alg}{\mathcal{A}}
\newcommand{\unlearn}{\mathcal{U}}

\newcommand{\dataset}{\mathcal{D}}
\newcommand{\forgetset}{\mathcal{S}}
\newcommand{\retainset}{\mathcal{D} \setminus \mathcal{S}}
\newcommand{\nanmax}{{\texttt{nan-max}}}

\usepackage{caption} 
\title{Are we making progress in unlearning? Findings from the first NeurIPS unlearning competition}

\author{%
Eleni Triantafillou$^1$\thanks{corresponding author: etriantafillou@google.com}
  \And
Peter Kairouz$^2$
\And
Fabian Pedregosa$^1$
\And
Jamie Hayes$^1$
\And
Meghdad Kurmanji$^3$
\And
Kairan Zhao$^3$
\And
Vincent Dumoulin$^1$
\And
Julio Jacques Junior$^{5,6}$
\And
Ioannis Mitliagkas$^{1,7}$
\And
Jun Wan$^{4,8}$
\And
Lisheng Sun Hosoya$^{4,9}$
\And
Sergio Escalera$^{4,5,6}$
\And
Gintare Karolina Dziugaite$^1$
\And
Peter Triantafillou$^3$
\And
Isabelle Guyon$^{1,4,9}$  
\AND \\
$^1$Google DeepMind \quad $^2$Google Research \quad $^3$University of Warwick \quad $^{4}$ChaLearn \\
\quad $^5$University of Barcelona \quad $^6$Computer Vision Center \quad $^7$University of Montreal\\
$^8$Institute of Automation, Chinese Academy of Sciences \quad $^9$Université Paris Saclay\\
}

\begin{document}

\maketitle

\begin{abstract}
We present the findings of the first NeurIPS competition on unlearning, which sought to stimulate the development of novel algorithms and initiate discussions on formal and robust evaluation methodologies. The competition was highly successful: nearly 1,200 teams from across the world participated, and a wealth of novel, imaginative solutions with different characteristics were contributed. In this paper, we analyze top solutions and delve into discussions on benchmarking unlearning, which itself is a research problem. The evaluation methodology we developed for the competition measures forgetting quality according to a formal notion of unlearning, while incorporating model utility for a holistic evaluation. We analyze the effectiveness of different instantiations of this evaluation framework vis-a-vis the associated compute cost, and discuss implications for standardizing evaluation. We find that the ranking of leading methods remains stable under several variations of this framework, pointing to avenues for reducing the cost of evaluation. Overall, our findings indicate progress in unlearning, with top-performing competition entries surpassing existing algorithms under our evaluation framework. We analyze trade-offs made by different algorithms and strengths or weaknesses in terms of generalizability to new datasets, paving the way for advancing both benchmarking and algorithm development in this important area.
\end{abstract}


\section{Introduction}

The trend of increasingly large and data-hungry deep learning models has led to exciting success stories. 
However, the heavy reliance of these models on training data has also generated important concerns.
These include legal, privacy, safety violations and inaccurate predictions, stemming from the perpetuation of harmful, incorrect or outdated training data.
A naive approach for correcting these issues is to simply remove the offending or no longer permissible subset of the training set and retrain ``from scratch''.
However, these increasingly large models are also increasingly expensive to train, making it impractical to retrain from scratch whenever a new problematic training data subset is identified.
We are thus faced with important technical challenges in designing machine learning pipelines that perform strongly, while allowing to efficiently comply with deletion requests.

Machine unlearning \citep{nguyen2022survey} has emerged as a research area to address this issue of efficiently erasing (the influence of) a subset of training data from a trained model. This is a challenging task, especially given the non-convex loss landscape of deep neural networks, where tracing the influence of a subset of training data on the model's weights and / or outputs, both accurately and efficiently, is an open problem \citep{koh2017understanding,bae2022if,barshan2020relatif,feldman2020does,paul2021deep,attias2024information}. Furthermore, imperfect attempts at erasing information from models may lead to sacrificing the utility of the model and its knowledge of permissible information. There are therefore complex trade-offs between \emph{forgetting quality, model utility} and \emph{efficiency} that further complicate the quest of designing practical unlearning algorithms and their evaluation.

While unlearning has gained increased attention, we argue that progress is significantly hampered by the challenge of designing benchmarks and operationally meaningful evaluation metrics. In particular, evaluation of unlearning is a research problem in and of itself: if measuring the influence of training data on models is an open problem, then so is measuring the remaining influence after unlearning. In organizing the NeurIPS'23 competition on unlearning, we had two key objectives: i) increase the visibility of this important problem and foster the creation of better unlearning algorithms, and ii) initiate a dialogue on rigorous evaluation methods by introducing a principled evaluation framework.

The NeurIPS'23 competition on unlearning\footnote{\url{https://unlearning-challenge.github.io/}} was designed to target a realistic scenario where an age predictor is trained on facial images and subsequently, a subset of users whose images were included in the training process request their data be deleted. The competition's participants were tasked with the goal of developing algorithms capable of erasing the influence of the data of those users from the model, without (overly) hurting its utility. The competition was hosted on Kaggle\footnote{\url{https://www.kaggle.com/competitions/neurips-2023-machine-unlearning/}} from September 11 to November 30, 2023. 
A total of 1,338 individuals from 72 countries participated. 
At the end of the competition, there were 1,121 teams and 1,923 submissions.

With nearly 1,200 teams participating, and a wide range of different algorithms proposed, we consider our first objective a resounding success. At the same time, the plethora of novel algorithms along with a multitude of pre-existing state-of-the-art unlearning algorithms presents us with a lot of exciting work for deepening our understanding of unlearning in general: What are algorithmic success and failure modes? 
Do the algorithms contributed in the competition outperform state-of-the-art unlearning algorithms? How does evaluation of these methods according to our metric agree or disagree with findings in the literature? All these lead to a key question: Are we making progress in machine unlearning? In this report, we seek to answer these questions through an extensive empirical evaluation and analyses. 

The principal contributions of this paper include:
an operational definition of unlearning that allows us to introduce a practical evaluation framework, an analysis of top-performing solutions from the first NeurIPS competition on unlearning, a discussion of the effectiveness of different instantiations of the evaluation framework concerning computational cost and of algorithm trade-offs in terms of utility and forgetting, as well as ease of generalizability to new datasets.


\section{Background} \label{sec:background}

\subsection{Defining machine unlearning}
Let $\theta^o = \alg(\dataset)$ denote the weights of a model (the ``original model'') obtained by applying learning algorithm $\alg$ on dataset $\dataset$. Informally, the goal of machine unlearning is to remove the influence of a forget set $\forgetset \subseteq \dataset$ from $\theta^o$. A straightforward solution is to simply retrain the model from scratch on an adjusted training set that excludes $\forgetset$, referred to as the ``retain set''. We denote by $\theta^r$ the weights of this ideal solution $\theta^r = \alg(\retainset)$. Unfortunately, retraining from scratch is inefficient and, in some cases prohibitively costly, depending on the size of the model and the frequency of unlearning requests. Therefore, instead of throwing away $\theta^o$ and retraining a new model, we seek an efficient unlearning algorithm $\unlearn$ that starts from $\theta^o$ and produces an unlearned model $\theta^u$ by post-processing: $\theta^u = \unlearn(\theta^o, \forgetset, \dataset)$. Intuitively, the ``closer'' $\theta^u$ is to $\theta^r$, the more successful $\unlearn$ is at unlearning. 

Measuring success of unlearning then requires estimating how close two distributions are to one another: the distribution of $\theta^u$ and that of $\theta^r$.  We refer to distributions here since running $\alg$ and $\unlearn$ with different random seeds that control, for instance, the initialization and order of mini-batches, will yield slightly different model weights each time. There are many approaches for estimating closeness of distributions, such as instantiating a Kolmogorov–Smirnov test or measuring the Kullback–Leibler divergence. We consider approaches that can be operationalized by mounting ``attacks'' that attempt to tell apart the two distributions and measuring closeness based on the degree of failure of the attack.

We now formalize the above intuition in a definition that is largely\footnote{Ours is a weaker notion that fixes the dataset and forget set, to precisely capture the competition setup.} inspired by \citep{sekhari2021remember, gupta2021adaptive, neel2021descent}, which in turn draw inspiration from differential privacy \citep{dwork2006differential}.
\begin{definition}{\bf $(\varepsilon, \delta)$-unlearning.} 
\label{defn:unlearning}
For a fixed dataset $\dataset$, forget set $\forgetset \subseteq \dataset$, and a randomized learning algorithm $\alg$, an unlearning algorithm $\unlearn$ is $(\varepsilon,\delta)$-unlearning with respect to $(\dataset, \forgetset, \alg)$ if for all $R \subseteq \mathcal{R}$ where $\mathcal{R}$ denotes the output space, in this case the space of model parameters $\theta$, we have:
\begin{align*}
\Pr[\alg(\retainset) \in R] &\le e^\varepsilon \Pr[\unlearn(\alg(\dataset), \forgetset, \dataset) \in R] + \delta,    \quad \mathrm{and} \\
\Pr[\unlearn(\alg(\dataset), \forgetset, \dataset) \in R]  &\le e^\varepsilon \Pr[\alg(\retainset) \in R] + \delta.     
\end{align*}
\end{definition}

The above definition expresses the degree of success of an unlearning algorithm $\unlearn$ (with respect to $\dataset$, $\forgetset$ and $\alg$) as a function of a notion of divergence between the distributions of $\theta^r$ and $\theta^u$. 

Specifically, the degree of success of unlearning is captured in the $\varepsilon$ and $\delta$ parameters. Notice that when $\varepsilon$ and $\delta$ are very small, the distributions of the retrained model $\theta^r = \alg(\retainset)$ and the unlearned model $\theta^u = \unlearn(\alg(\dataset), \forgetset, \dataset)$ are nearly indistinguishable from one another, signalling successful unlearning. In this work, we will compare unlearning algorithms to one another by fixing $\delta$ to a small value, and computing each algorithm's $\varepsilon$. For a pair of algorithms $\unlearn_1$, $\unlearn_2$, we will say that $\unlearn_1$ is better according to this metric than $\unlearn_2$ (with respect to $\dataset$, $\forgetset$ and $\alg$) if it yields a smaller $\varepsilon$ than that of $\unlearn_2$.

We refer to an unlearning algorithm $\unlearn$ that is $(0,0)$-unlearning as \textit{exact unlearning}. For non-convex models, the only known approach to exact unlearning involves retraining from scratch (parts of) the model. This can be done either naively or through cleverly-designed mixture models, where one only needs to retrain the component(s) affected by an unlearning request \citep{bourtoule2021machine}. In the worst case, though, where a forget set is distributed across the training sets of all sub-models, even clever systems suffer the same computational cost as naive retraining, and these mixture models may also have poorer utility compared to other architectures. Motivated by these challenges, the community has recently developed a plethora of \textit{approximate unlearning} methods, whose $\varepsilon$ and $\delta$ values are generally not known theoretically but are significantly more computationally efficient or perform better than exact methods according to empirical metrics. The competition focused on approximate unlearning.

\subsection{Empirical estimation of \texorpdfstring{$\varepsilon$}{eps} via a hypothesis-testing interpretation}
Drawing inspiration from empirical estimation of $\varepsilon$ for differential privacy (DP), we describe an evaluation procedure for unlearning that can be interpreted as a hypothesis test: the null hypothesis is that $\alg$ was trained on all of $\dataset$, and unlearning was then applied to erase $\forgetset$, and the alternative hypothesis is that $\alg$ was trained on $\retainset$. False positives (type-I errors) occur when the null hypothesis is true but is rejected and false negatives (type-II errors) when the alternative hypothesis is true but is rejected. We borrow the result of \cite{kairouz2015composition} that characterizes $(\varepsilon,\delta)$-DP in terms of the false positive rate (FPR) and false negative rate (FNR) achievable by an acceptance region, and use it for empirical estimation of $\varepsilon$ for unlearning at any fixed $\delta$. 

\begin{namedtheorem}[Theorem 2.2 (adapted from \cite{kairouz2015composition})]
\label{thrm:hyp_testing}
Fix $\dataset$, $\forgetset \subseteq \dataset$, and a randomized learning algorithm $\alg$. Assume $\unlearn$ is an $(\varepsilon,\delta)$-unlearning algorithm with respect to $(\dataset, \forgetset, \alg)$. Let 
 $X$ be sampled from either $\alg(\retainset)$  or $\unlearn(\alg(\dataset), \forgetset, \dataset)$.  Then the performance of any (possibly randomized) hypothesis testing rule that tries to distinguish whether $X$ came from  $\alg(\retainset)$ or   $\unlearn(\alg(\dataset), \forgetset, \dataset)$ is governed by
\begin{align*}
\text{FPR} + e^\varepsilon \text{FNR} &\geq   1 - \delta,    \quad \mathrm{and} \\
\text{FNR} + e^\varepsilon \text{FPR} &\geq   1 - \delta.     
\end{align*}
\end{namedtheorem}

Intuitively, if $\varepsilon$ and $\delta$ are both small, the above theorem states that regardless of computation power of the testing rule, it is statistically impossible to get FPR and FNR to be simultaneously small. This characterization enables estimating $\varepsilon$ at a fixed $\delta$ as

\begin{equation} \label{eq:eps-from-fpr-fnr}
\hat{\varepsilon} = \max\left\{\log \frac{1-\delta- \hat{\mbox{FPR}}}{\hat{\mbox{FNR}}}, \log \frac{1-\delta-\hat{\mbox{FNR}}}{\hat{\mbox{FPR}}}\right\},
\end{equation}

where $\hat{\mbox{FPR}}$ and $\hat{\mbox{FNR}}$ are estimates of the true FPR and FNR under an instantiated \textit{attack} that is designed to predict as accurately as possible whether a given model was obtained through recipe $\unlearn(\alg(\dataset), \forgetset, \dataset)$ or $\alg(\retainset)$. Such a prediction problem is closely related to \emph{membership inference attacks} that aim to infer whether a particular example was included in the training set of a given model \citep{shokri2017membership, jagielski2020auditing}.
Recent work designs adaptations of such attacks for evaluating unlearning \citep{kurmanji2024towards,pawelczyk2023context,hayes2024inexact}.

\section{Introducing our evaluation framework for unlearning}

Given the challenges around constructing unlearning algorithms that provably satisfy the $(\varepsilon, \delta)$-unlearning notion given in Definition \ref{defn:unlearning}, we now devise a principled  methodology for empirically evaluating and ranking  unlearning algorithms. 

Evaluating an unlearning algorithm $\unlearn$ requires carefully quantifying three key components: (a) the ``forgetting quality'' (denoted by $\mathcal{F}$) of $\unlearn$, (b) the utility of $\theta^u$, and (c) the efficiency of running $\unlearn$.  We can measure the last two via standard, existing metrics. Specifically, we measure utility through accuracy on the retain and test sets, and we ensure that all algorithms that we study are substantially more efficient than retraining by enforcing a hard cut-off on the runtime of unlearning (roughly 20\% of the time it takes to retrain-from-scratch). Defining $\mathcal{F}$ is involved and is deferred to Section \ref{sec:forget_quality}.Our final score combines an estimate of $\mathcal{F}$,
where higher is better, with an estimate of utility:
\begin{align} \label{eq:scores}
\text{Final score} = \mathcal{F} \times \frac{\text{Acc}(\retainset, \theta^u)}{\text{Acc}(\retainset, \theta^r)} \times \frac{\text{Acc}(\dataset_{test}, \theta^u)}{\text{Acc}(\dataset_{test}, \theta^r)}, &&
\text{Acc}(\dataset, \theta) = \frac{1}{|\dataset|}\sum_{(x, y) \in \dataset} [f(x; \theta) = y],
\end{align}
where $f(x;\theta)$ is the network function, parameterized by $\theta$, and mapping input $x$ to a class.

Intuitively, the above formula adjusts the forgetting quality $\mathcal{F}$ to take utility in consideration, by penalizing an unlearning algorithm if either its retain or test accuracy is smaller than the corresponding accuracy of retraining-from-scratch. Next, we dive in to discussing how we obtain $\mathcal{F}$.

\subsection{Measuring forgetting quality \texorpdfstring{$\mathcal{F}$}{F} via empirical estimation of \texorpdfstring{$\varepsilon$}{eps}} \label{sec:forget_quality}

A straightforward way to measure forgetting quality is via the hypothesis testing interpretation of $(\varepsilon, \delta)$-unlearning. Specifically, $\varepsilon$ 
is estimated by instantiating various attacks that directly attempt to distinguish between the distributions of $\theta^r$ and $\theta^u$. However, this approach faces two important challenges: (a) $\theta^r$ and $\theta^u$ can be high dimensional, leading to a computational difficulty around running the attacks directly on their parameters, and (b) two unlearning algorithms $\unlearn_1$, $\unlearn_2$ may have the same (or similar) estimated $\varepsilon$ overall, but could still behave very differently on many examples $s \in \forgetset$ (as we show in Section \ref{sec:per_example_breakdown}), requiring a more granular resolution for comparing $\unlearn_1$ to $\unlearn_2$.

To address the first challenge, we will consider (in Section \ref{sec:eps_computation}) distributions of (scalar) outputs of unlearned and retrained models when given as input examples from $\forgetset$, rather than distributions in weight space \footnote{Prior work finds that white-box access does not necessarily always translate to an improved attack~\citep{sablayrolles2019white}.}. To address the second challenge, we will estimate the $\varepsilon$ for each $s \in \forgetset$ and then provide (in Section \ref{subsub:aggregate_eps}) a methodology for binning and aggregating the estimated $\varepsilon$ to obtain $\mathcal{F}$. 

\begin{figure}[t!]
\begin{subfigure}[b]{0.49\textwidth}
     \centering
    \includegraphics[scale=0.115]{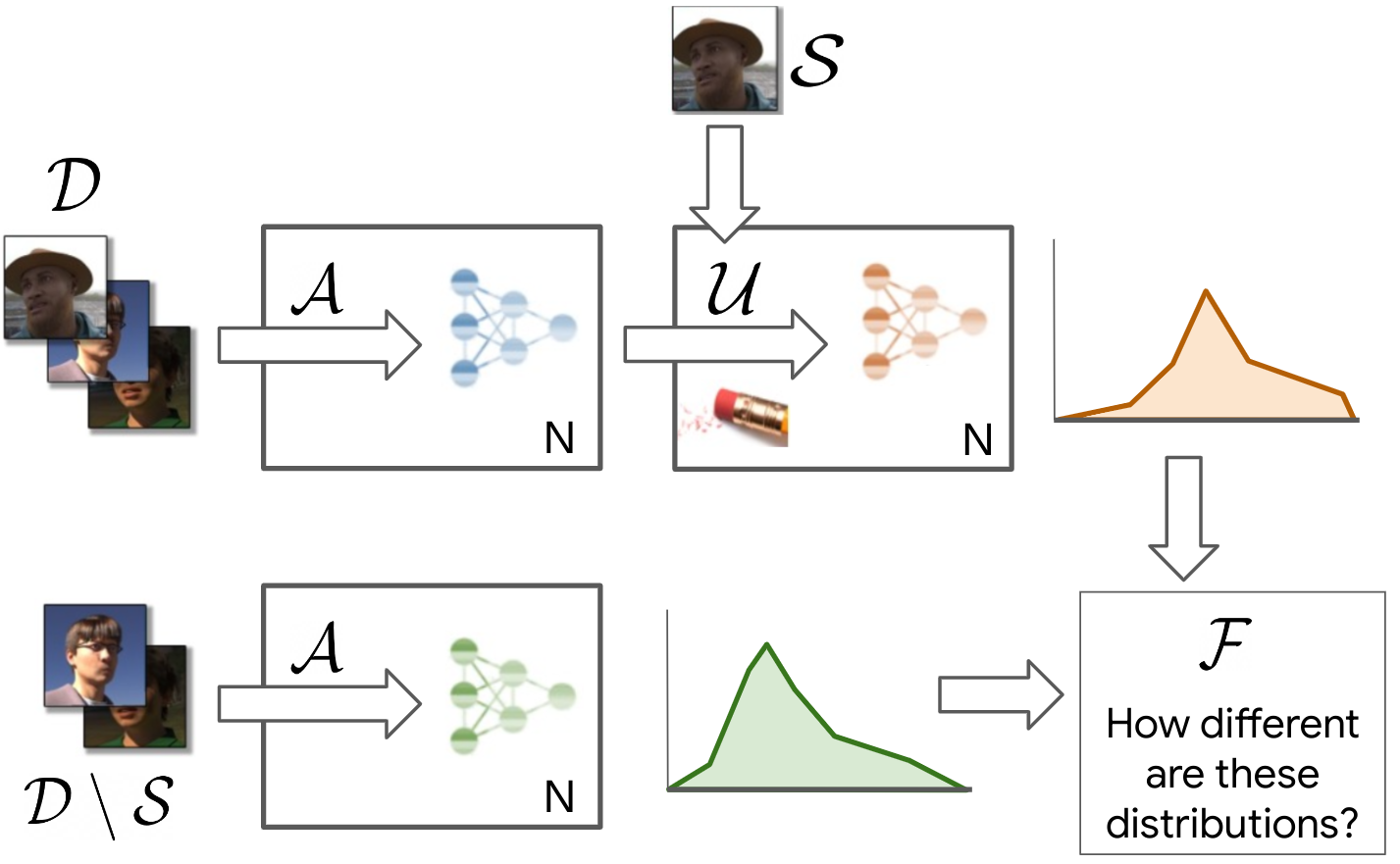}
\end{subfigure}
\begin{subfigure}[b]{0.49\textwidth}
     \centering
    \includegraphics[scale=0.14]{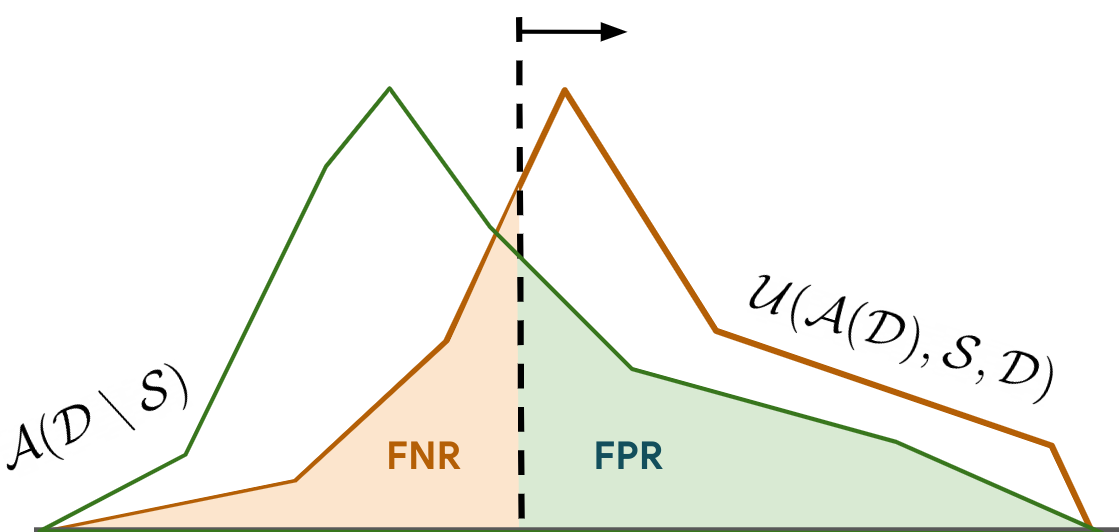}
\end{subfigure}
\caption{\textbf{Left}: overview of the evaluation of forgetting quality. We draw $N$ samples of $\theta^o$ and $\theta^r$, by repeating procedures $\alg(\dataset)$ and $\alg(\dataset \setminus \forgetset)$, respectively, $N$ times, with different random seeds. Then, we obtain $N$ samples of $\theta^u$ by applying $\unlearn$ on each of the original model samples $\theta^o$. We then compute an estimate of forgetting quality $\mathcal{F}$ based on how similar the distributions of $\theta^u$ and $\theta^r$ are, according to a 1-dimensional test statistic (Section \ref{sec:eps_computation}). Closeness of those distributions indicates good unlearning, associated with higher $\mathcal{F}$-score. \textbf{Right}: example decision rule to separate the histograms of a 1-dimensional test statistic of the distributions of $\theta^u$ and $\theta^r$, for a given example in the forget set (see Section \ref{sec:eps_computation}). This decision rule predicts ``unlearned'' for values greater than the threshold shown as a black dotted line. As we describe in Section \ref{sec:eps_computation}, we sweep several thresholds and use the one that best separates the two distributions to measure their closeness.}
\label{fig:overview} 
\end{figure}

Putting these two observations together, our procedure for obtaining $\mathcal{F}$ has two steps: i) For each $s \in \forgetset$, estimate the $\varepsilon$ for the distributions of outputs of unlearned and retrained models when given as input $s$, and ii) aggregate the per-example $\varepsilon$'s to get an overall estimate of forgetting quality $\mathcal{F}$. Overall, this amounts to estimating the discrepancy between the outputs of $\theta^u$ and $\theta^r$ across all of $\forgetset$.  We overview evaluation of forgetting quality in Figure \ref{fig:overview}.



\subsubsection{Computing \texorpdfstring{$\varepsilon$}{eps} for a given forget set example} \label{sec:eps_computation}
For an example $s \in \forgetset$, we denote the outputs under unlearned and retrained models as $h(f(s; \theta^u))$ and $h(f(s; \theta^r))$, respectively, where $h$ is the function used in \citep{carlini2022membership} that applies logit-scaling on the model's ``confidence'', i.e. a scalar corresponding to the probability of the correct class. The logit-scaling is designed to make the distribution of that scalar more ``Gaussian''. We begin by empirically estimating the distributions of $h(f(s; \theta^u))$ and $h(f(s; \theta^r))$ via $N$ samples of each. This is done by using $N$ samples from each of $\theta^u = \unlearn(\alg(\dataset), \forgetset, \dataset)$ and $\theta^r = \alg(\retainset)$ (corresponding to unlearned and retrained models, respectively), forward-passing example $s$ through each of these models, and post-processing the outputs with $h$. This leaves us with $N$ samples from each of $h(f(s; \theta^u))$ and $h(f(s; \theta^r))$, each of which is a scalar.

To leverage Equation \ref{eq:eps-from-fpr-fnr} to compute $\varepsilon$, we require estimates of $\hat{\mbox{FPR}}$ and $\hat{\mbox{FNR}}$ that characterize the degree of success of an attacker in separating the two distributions; the attacker is better the smaller $\hat{\mbox{FPR}}$ and $\hat{\mbox{FNR}}$ are, signalling poorer unlearning. However, we are not interested in the $\hat{\mbox{FPR}}$ and $\hat{\mbox{FNR}}$ of \textit{any} attacker: the failure of a poor attacker at distinguishing the two distributions does not say much about the quality of unlearning. Therefore, we design a procedure that aims to get the strongest attacker for each example (within a parameterized family of computationally tractable attackers), and we use its $\hat{\mbox{FPR}}$ and $\hat{\mbox{FNR}}$ estimates to compute $\varepsilon$. Practically, we consider $m$ ``attacks'', compute the $\hat{\mbox{FPR}}$ and $\hat{\mbox{FNR}}$ of each and associated estimates of $\varepsilon$ and we keep the largest out of those $\varepsilon$'s; i.e.\ the one corresponding to the strongest attack. The intuition is that, for unlearning to be successful, it must defend against the strongest attack. We outline this procedure in Algorithm \ref{alg:eps-from-fpr-fnr}.


More concretely, we consider two families of attacks: i) single-threshold and ii) double-threshold attacks. In the first family, each attack is instantiated by the choice of a threshold $t$ and the associated decision rule is that any value $> t$ is predicted to belong to whichever of the two distributions (unlearned or retrained) has a larger median. In the second family, each attack is instantiated by the choice of two thresholds $t_1$ and $t_2$ with $t_1 < t_2$, with any values in between $t_1$ and $t_2$ being predicted as belonging to the peakiest distribution. We create many instantiations of each of these two families of attacks (by sweeping several values for $t$, $t_1$ and $t_2$), compute the $\hat{\mbox{FPR}}$ and $\hat{\mbox{FNR}}$ of each, and as described above and in Algorithm \ref{alg:eps-from-fpr-fnr}, we keep the $\varepsilon$ of the strongest attack across both families. We note that, each decision rule can simply be regarded as a classifier (linear, in the case of single-threshold and non-linear in the case of double-threshold) for separating the outputs of the unlearned and retrained models, for a given example.

Including double-threshold attacks is important as we found that unlearning algorithms produce different forms of distributions, some more peaky than others, and not all are easily separable from the retrain-from-scratch solution using only one threshold (see Section \ref{sec:histogram_visualizations} for some examples). We further discuss this issue, and other important ``implementation details'' in Section \ref{sec:eval_in_detail}.

\begin{figure}[t!]
\begin{subfigure}[b]{0.33\textwidth}
     \centering
    \includegraphics[scale=0.08]{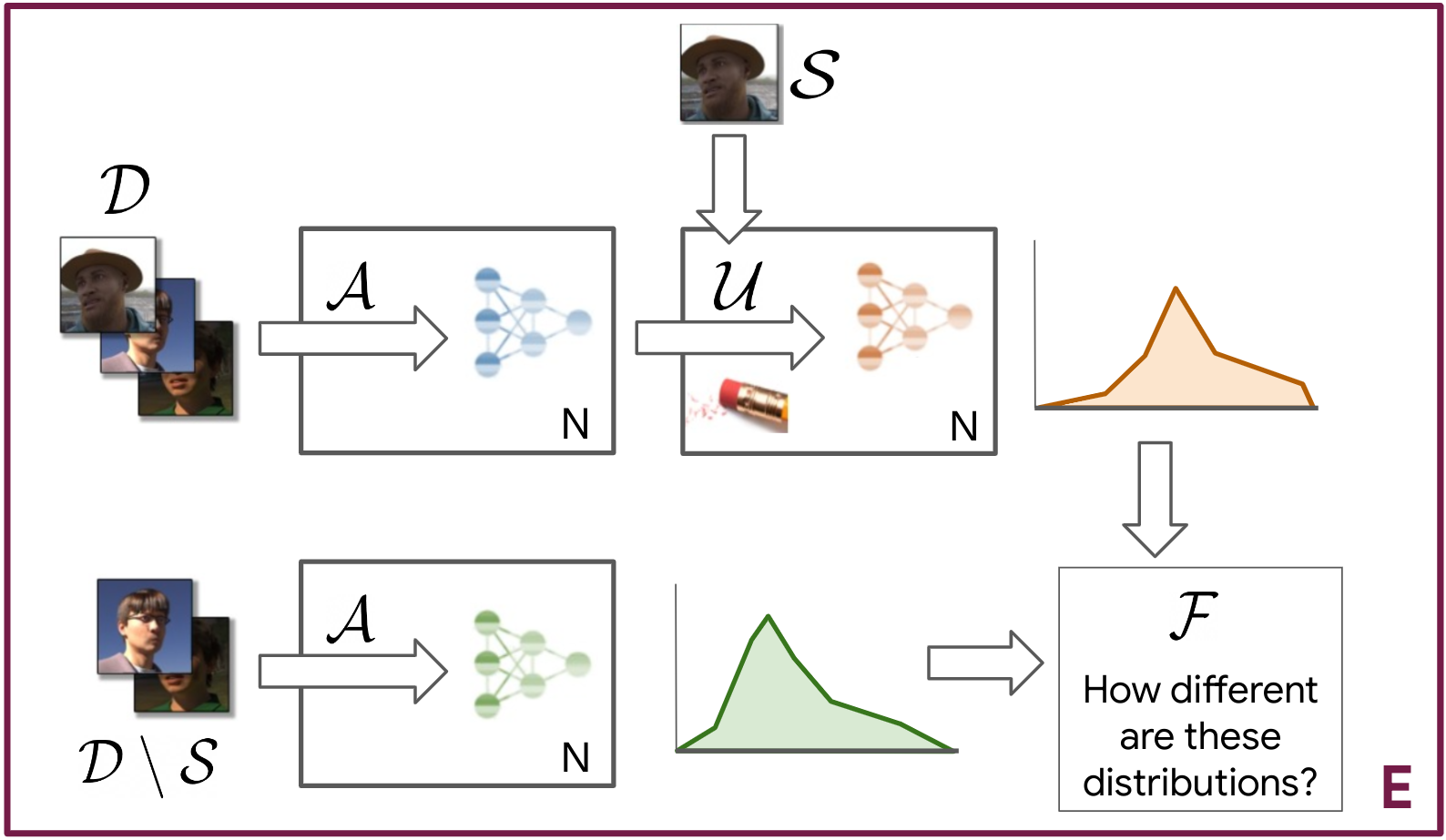}
\end{subfigure}
\begin{subfigure}[b]{0.33\textwidth}
     \centering
    \includegraphics[scale=0.08]{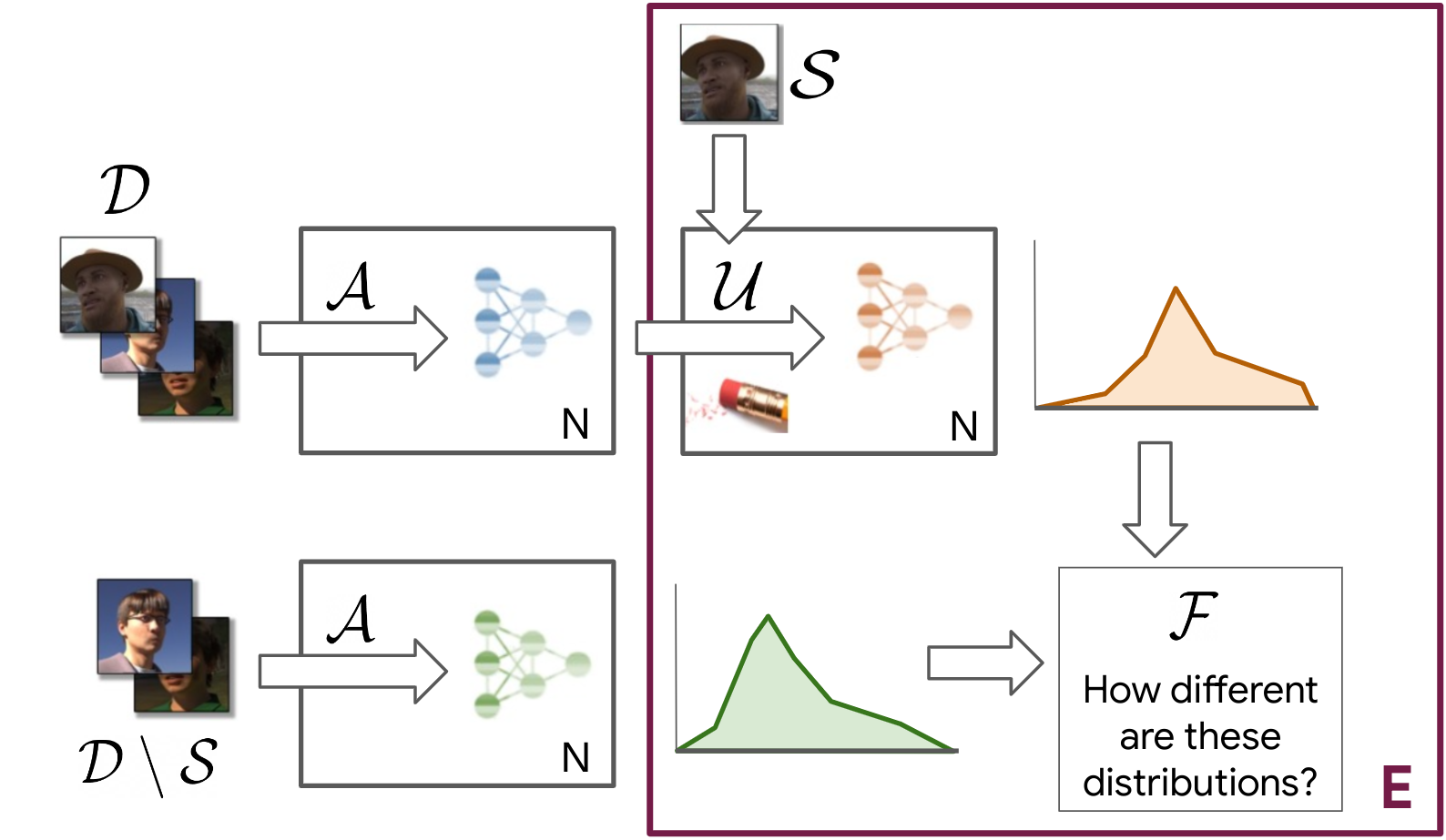}
\end{subfigure}
\begin{subfigure}[b]{0.33\textwidth}
     \centering
    \includegraphics[scale=0.08]{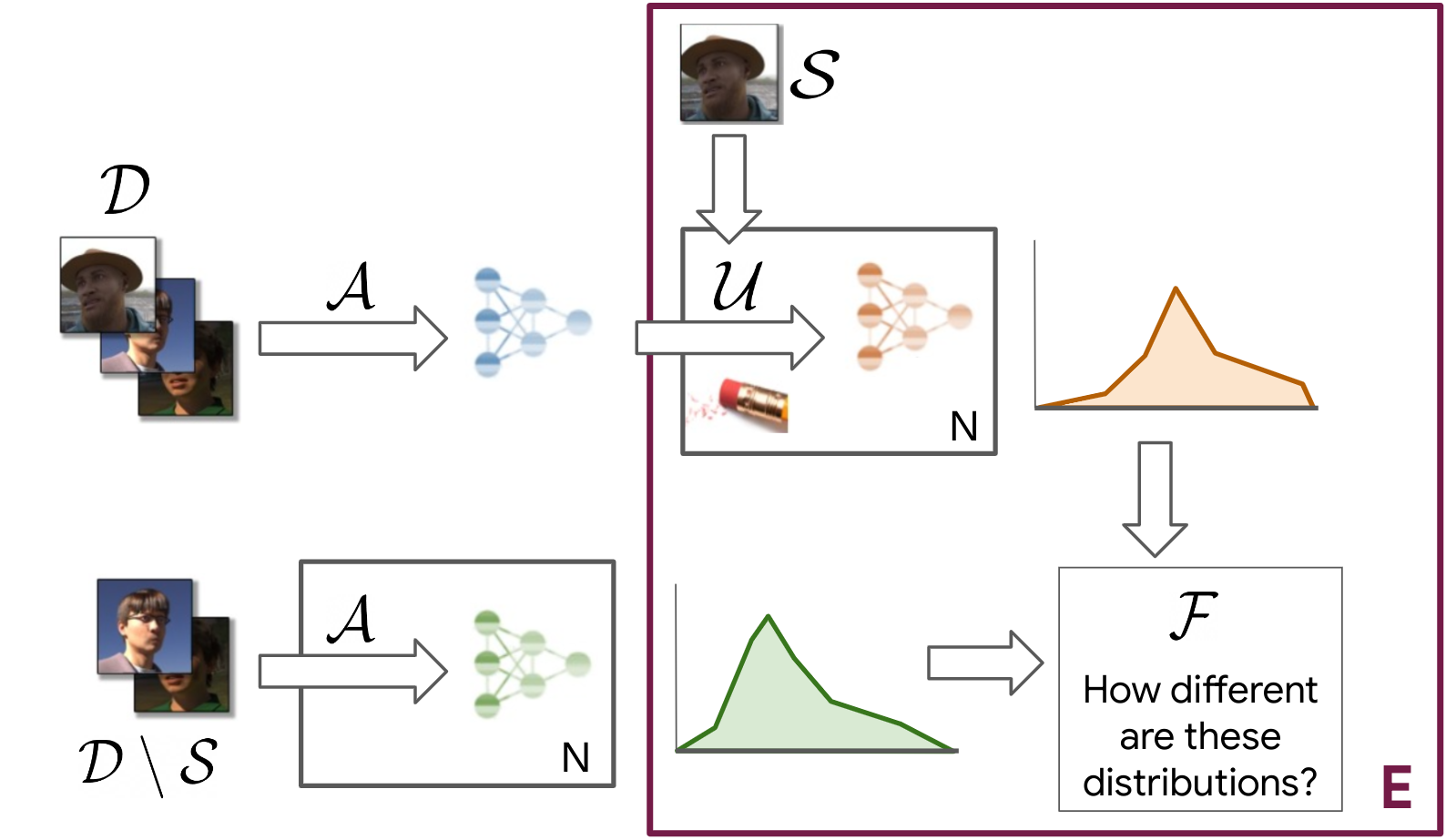}
\end{subfigure}
\caption{Practical instantiations of our evaluation framework that explore the accuracy / efficiency trade-off. In each case, $N$ samples from each of $\theta^u$ and $\theta^r$ are used to compute an estimate of $\mathcal{F}$, and we obtain $E$ samples of $\mathcal{F}$ to compute confidence intervals. Each setup differs in how much ``work'' is reused across the $E$ ``experiments''. The plate notation (a rectangle with a number at its bottom right corner (e.g. $N$ or $E$)) denotes that the contents of the rectangle are repeated that number of times.
\textbf{Left, Setup ``Full''}: In each of $E$ experiments, we draw $N$ samples of every distribution. This is the statistically correct variant as it yields $E$ i.i.d samples of $\mathcal{F}$ but is very costly.
\textbf{Middle, Setup ``Reuse-$N$-$N$''}: $N$ samples of each of $\theta^o$ and $\theta^r$ are drawn once and reused across the $E$ runs, each of which simply runs $\mathcal{U}$ on top of each sample of $\theta^o$. \textbf{Right, Setup ``Reuse-$N$-1''}: a single sample of $\theta^o$ is used to obtain all samples of $\theta^u$, and a single set of $N$ samples of $\theta^r$ is reused for all $E$ runs.}
\label{fig:setups}
\end{figure}
\subsubsection{Aggregating per-example \texorpdfstring{$\varepsilon$}{eps}'s to obtain an overall estimate of forgetting quality} \label{subsub:aggregate_eps} We now discuss how to use per-example $\varepsilon$'s to obtain the overall estimate of forgetting quality $\mathcal{F}$. To that end, we define a scoring function $\mathcal{H}$ that awards a number of ``points'' for each example, based on that example's $\varepsilon$, and we aggregate across $\forgetset$ by averaging the per-example scores. Specifically:
\begin{align*}
    \mathcal{F} = \frac{1}{|\mathcal{S}|} \sum_{s \in \mathcal{S}} \mathcal{H}(\varepsilon^s), && \mathcal{H}(\varepsilon) = \frac{2}{2^{n(\varepsilon)}}, && n(\varepsilon) = \text{floor}\left(\frac{\varepsilon}{\texttt{bin-width}}\right),
\end{align*}
where $n$ is a function that maps an $\varepsilon$ to a ``bin index'' (an integer in the range $[1, B]$, where $B$ is the total number of bins). We set \texttt{bin-width} to 0.5. Notice that the smaller the $\varepsilon$ for an example (indicating better unlearning of that example), the smaller $n(\varepsilon)$ is, and thus the more points will be awarded by $\mathcal{H}$. By aggregating over $\mathcal{H}$-scores then, $\mathcal{F}$ offers an overall estimate of how well $\mathcal{S}$ was unlearned, where higher is better. We chose to use binning as it is more granular than a simple average or computing quantiles and less sensitive to noise compared to directly using estimated $\varepsilon$'s. 




\section{Instantiating our framework: the accuracy / efficiency divide} \label{sec:practical_instantiations}

We now present practical instantiations of our framework offer differing accuracy-efficiency trade-offs. This is key as accurate evaluation is computationally expensive and, at the same time, it is imperative to produce confidence intervals and remove statistical dependencies when computing our statistic.

More concretely, the procedure we outlined above requires sampling from three distributions: the distribution of i) $\theta^r = \alg(\retainset)$,  ii) $\theta^u = \unlearn(\alg(\dataset), \forgetset, \dataset)$, which in turn requires estimating the distribution of iii) $\theta^o = \alg(\dataset)$. Drawing a sample from each requires running $\alg$ (for i and iii) and $\unlearn$ (for ii). Let $N$ denote the number of samples from each of $\theta^u$ and $\theta^r$ that are fed into our evaluation framework to compute $\mathcal{F}$, and let $E$ denote a number of ``experiments'', each of which produces an estimate of $\mathcal{F}$. We use $E > 1$ to produce confidence intervals over $\mathcal{F}$.

While running $\unlearn$ is relatively inexpensive (we consider unlearning algorithms that are much faster than training from scratch), running $\alg$ is costly. Based on this observation, we consider three setups offering different trade-offs between accuracy and compute cost. The gold standard is \textbf{Setup ``Full''}, where each of $E$ experiments draws $N$ fresh samples of every distribution. This is the most statistically correct variant, as it leads to $E$ i.i.d.\ samples of the forgetting quality $\mathcal{F}$, used to compute confidence intervals. However, it is by far the most computationally intensive. Significantly saving on compute, \textbf{Setup ``Reuse-$N$-$N$''} draws $N$ samples form each of $\theta^o$ and $\theta^r$ once and reuses them across experiments. Each experiment entails running $\unlearn$ to convert each sample of $\theta^o$ into a sample of $\theta^u$, yielding $N$ in total per experiment. Finally, further simplifying, \textbf{Setup ``Reuse-$N$-$1$''} uses a single sample of $\theta^o$ to obtain all samples of $\theta^u$ (``reuse $1$'') and a single set of $N$ samples of $\theta^r$ are reused across all experiments (``reuse $N$''). We illustrate these in Figure \ref{fig:setups}. 




Orthogonally, we also explored using \textbf{bootstrapping} 
to reduce the computation cost of Setup ``Full''. Specifically, from a pool size of $K$ triplets of $(\theta^o, \theta^u, \theta^r)$, we sample $N$ of them with replacement, and compute an estimate of $\mathcal{F}$. We repeat this procedure $E$ times, yielding a total of $E$ estimates of $\mathcal{F}$, as in Setup ``Full'', but requiring only $K$ models from each distribution here for all $E$ estimates, rather than $N \times E$ as in Setup ``Full''. In both cases, we set $E$ to 20 in our experiments, unless otherwise specified. We experiment with different values of $K$ and $E$ and report additional results in Section \ref{sec:bootstrapping}; by default we use $K = N \times 8$; yielding substantial compute savings over Setup ``Full''.



\textbf{The competition setup:} The competition, hosted on Kaggle, targeted a realistic scenario where an age predictor is trained on facial images from the CASIA-SURF dataset \citep{zhang2020casia} and subsequently, a subset of users whose images were included in the training process request their data be ``forgotten''. Due to practical considerations, we used Setup ``Reuse-$N$-$1$'', with $N = 512$ and $E = 1$. We describe the competition setup in full details in Section \ref{sec:details_competition}.

In this report, we empirically investigate how different the estimates of $\mathcal{F}$ are under the different setups (see Table \ref{tab:compute_of_instantiations} for overview of the compute of each), including the cheapest setup, used in the competition (see Figure \ref{fig:setup_A_B_C}; see Section \ref{sec:effect_of_N} for analysis of the effect of $N$). We run experiments to analyze different practical instantiations, revealing paths for compute-efficient proxies of $\mathcal{F}$.

\vspace{-3mm}
\section{Unlearning methods}
\begin{figure}[tb!]
    \centering
    \includegraphics[width=0.9\textwidth]{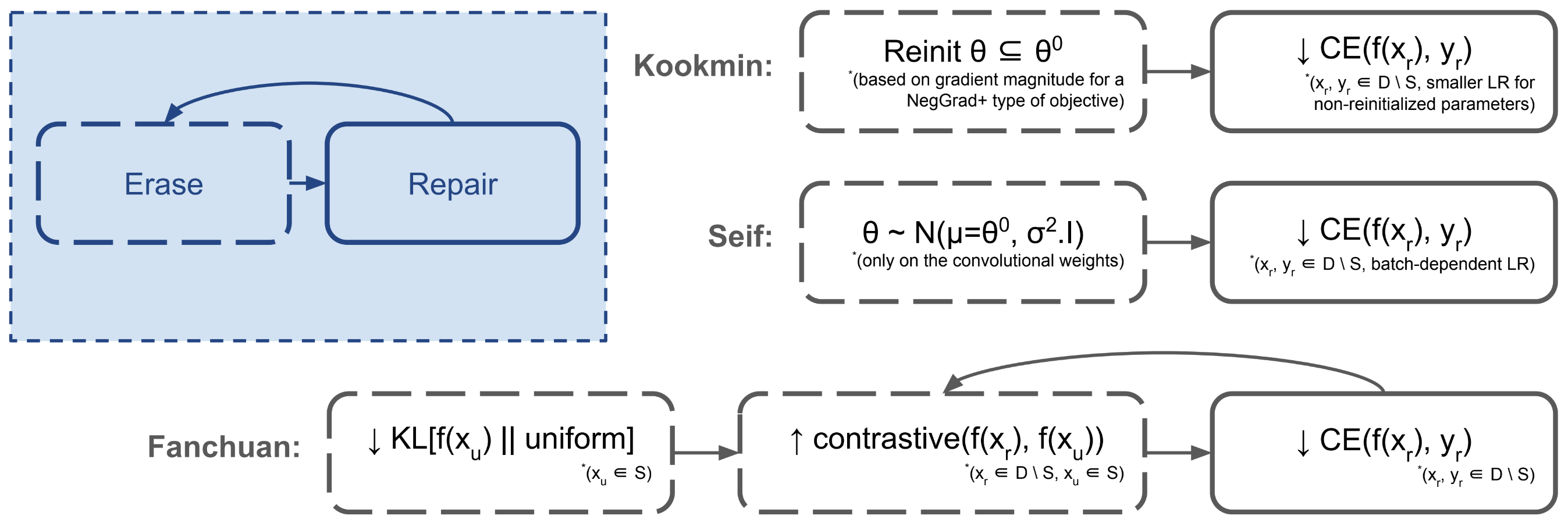}
    \caption{Commonalities between participants' methods. 
    We illustrate the top three approaches here and provide diagrams for all analyzed competition methods in \autoref{fig:methods_overview_full}.}
    \label{fig:methods_overview}
\end{figure}

\vspace{-2mm}
In this section, we describe the top methods from the competition that we analyze in this report\footnote{We chose methods that ranked 1--8 on the leaderboard, excluding 5th place which we couldn't reproduce.}, as well as state-of-the-art approaches that we compare against.
We observe that most submitted algorithms can be seen as being comprised of an ``erase'' phase, aiming to remove the influence of the forget set, followed by a ``repair'' phase, aiming to repair any excess damage to the utility of the model that is caused by imperfect erasing (\autoref{fig:methods_overview}).\footnote{Note that unlearning phases cannot always be neatly categorized as one of the two: for instance, finetuning on the retain set still has an indirect erasing effect on the forget set due to catastrophic forgetting.} See Section \ref{sec:detailed_algorithms} for more details.

Some methods implement the ``erase'' phase by reinitializing a subset of the layers, either heuristically (Amnesiacs, Sun), through random selection (Forget), or based on gradient (Kookmin) or parameter (Sebastian) norm. Other methods apply additive Gaussian noise to the parameters of a heuristically (Seif) or randomly (Sun) chosen subset of layers. Fanchuan implements two ``erase'' phases: the first pulls the model predictions for forget examples towards a uniform distribution, and the second attempts to maximize a dot-product contrastive loss between retain and forget predictions.

Most approaches implement the ``repair'' phase by directly minimizing a cross-entropy loss on the retain set (Fanchuan, Amnesiacs, Sun) while possibly modulating the learning rate based on a per-parameter (Kookmin, Sun) or per-batch (Seif) basis. The ``Sebastian'' method combines a weighted (reciprocal class weights) cross-entropy loss and a mean-squared error pulling the model's prediction entropy towards the original model's prediction entropy on retain examples. Amnesiacs implement two ``repair'' phases: the first minimizes the KL-divergence between the model's prediction and the original model's prediction on held-out validation examples, and the second combines the cross-entropy loss with a symmetric KL-divergence between the model's prediction and the original model's prediction on retain examples. The ``Forget'' method pulls the model's predictions towards the original model's predictions with a mean squared error loss for noisy retain examples.

We also compare against the Finetune baseline that simply finetunes the original model on the retain set, relying on ``catastrophic forgetting'' to remove the effect of the forget set, and five state-of-the-art algorithms: i) NegGrad+ and SCRUB \citep{kurmanji2024towards}, based on gradient ascent on the forget set and descent on the retain set, simultaneously, either using standard cross-entropy (NegGrad+) or distillation from the original model (SCRUB), ii) Random Label \citep{golatkar2020eternal,fan2023salun} that assigns a random label to each example in the forget set and finetunes on that re-labelled forget set; we added a ``repair phase'' that finetunes on the retain set, in line with other methods. iii) SalUn based on Random-Label \citep{fan2023salun}, that enhances that unlearning algorithm by only allowing it to update the ``salient weights'', determined based on the magnitude of gradients of a simple unlearning step (gradient ascent on the forget set).

\vspace{-2mm}
\section{Experimental investigation}
We now present experiments designed to answer: \textbf{Q1}: How do top algorithms from the competition compare to one another under practical instantiations of our framework? \textbf{Q2}: How do those algorithms fare against the state-of-the-art from the literature? \textbf{Q3}: How do they trade-off forgetting quality and utility? \textbf{Q4}: How does $\mathcal{F}$ correlate with other proxies for forgetting quality? \textbf{Q5}: How generalizable are different algorithms in terms of performance on another dataset after minimal-tuning?

We also perform analyses to examine the distributions of per-example $\varepsilon$ values and whether different algorithms find the same examples hard (Section \ref{sec:per_example_breakdown}), stitching together the ``erase'' and ``repair'' phases of different algorithms (Section \ref{sec:stitch_erase_repair}), investigating the relationship of forgetting quality estimates of $\mathcal{F}$ and a simple MIA \ref{sec:simple_mia}, measuring the potential effect of overfitting the attacker, due to choosing the (pair of) threshold(s) on the same set of $N$ samples from each distribution on which it is evaluated (Section \ref{sec:degree_of_overfitting}), measuring the effect of $N$ (Section \ref{sec:effect_of_N}) and $K$ (Section \ref{sec:bootstrapping}).

We conduct our investigation primarily (except for Q5) on CASIA-SURF, using setup ``Full'', $N$ = 1024, $E$ = 20 (see Section \ref{sec:practical_instantiations}) unless stated otherwise. See Section \ref{sec:implementation_details} for details.

\begin{figure}[ht]
\centering
\includegraphics[scale=0.24]{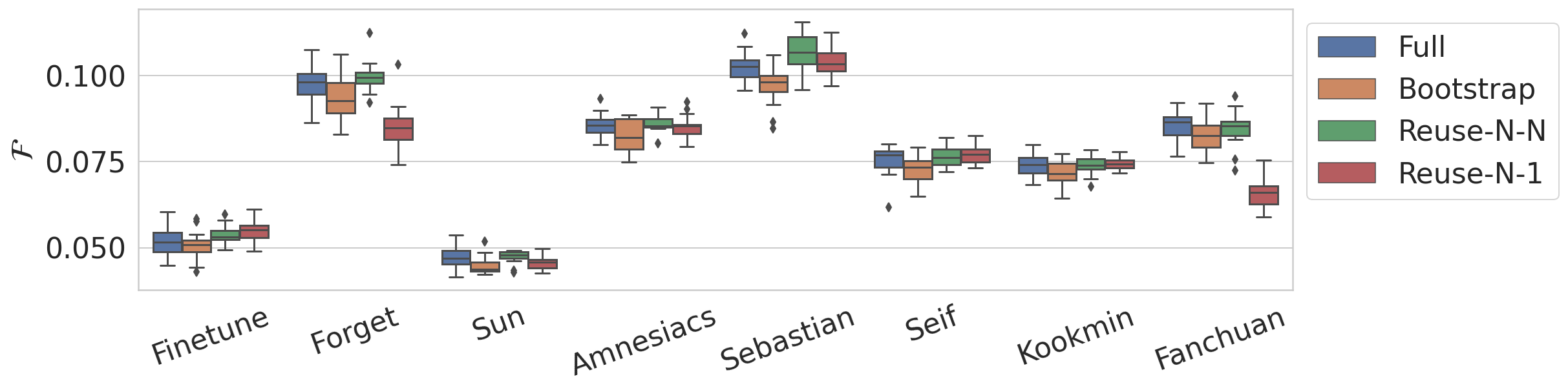}
\caption{$\mathcal{F}$-scores obtained by different setups trading-off accuracy / efficiency (see Section \ref{sec:practical_instantiations}). $N$ = 1024, $E$ = 20. 
}
\label{fig:setup_A_B_C}
\end{figure}

\textbf{Q1. Exploring practical instantiations of our framework} Figure \ref{fig:setup_A_B_C} compares algorithms under different setups (Section \ref{sec:practical_instantiations}), exploring the accuracy / efficiency trade-off of the evaluation method. We observe that Setup ``Reuse-$N$-$1$'' underestimates Fanchuan's $\mathcal{F}$-score. We note that this method is the most deterministic (it does not explicitly add noise), making it perhaps harder to cover the entire distribution of retrained models when starting from a single sample of $\theta^o$, especially if getting ``unlucky'' with the choice of that sample. However, aside from Setup ``Reuse-$N$-$1$'' (the cheapest one, which was used during the competition), the estimates of $\mathcal{F}$ are similar across remaining setups and the relative ranking of algorithms is stable, surfacing directions for mitigating the cost of evaluation.
Sections \ref{sec:degree_of_overfitting} and \ref{sec:effect_of_N} explore other variations of the evaluation that also preserve the ranking.

\begin{figure}[t]
\centering
\includegraphics[scale=0.23]{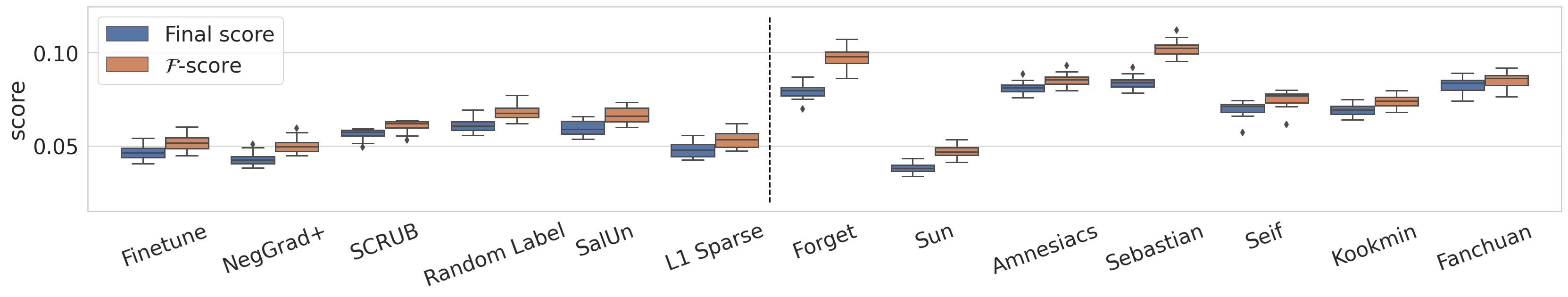}
\caption{Comparing leading competition algorithms (to the right of the dotted line) against state-of-the-art from the literature (to the left of the dotted line). We notice that several algorithms from the competition outperform existing ones according to our metrics. Setup ``Full'', $N$ = 1024, $E=10$.}
\label{fig:sota_comparison}
\end{figure}

\textbf{Q2. Comparison with the state-of-the-art} In Figure \ref{fig:sota_comparison}, we compare seven top submissions and five state-of-the-art methods. We observe that several top methods outperform existing ones, both in terms of their $\mathcal{F}$-score as well as the final score, after utility adjustment (Equation \ref{eq:scores}).


\textbf{Q3. Trade-off between forgetting and utility} Comparing the final score to the $\mathcal{F}$-score in Figure~\ref{fig:sota_comparison}, we observe that different methods differ in terms of their utility cost. Notably, the method with the best $\mathcal{F}$-score (Sebastian) is most penalized due to utility; a large drop that is perhaps expected since this methods prunes 99\% of the weights. We investigate the utility / forgetting quality trade-off further in Figure \ref{fig:utility-forgetting-tradeoff}. 
We notice that some unlearning methods harm utility more than others; some harming retain more than test accuracy, or vice versa. We discuss algorithms' trade-off profiles in Section \ref{sec:utility_forgetting_quality}.

\begin{wrapfigure}{r}{0.6\textwidth}
\centering
\includegraphics[scale=0.24]{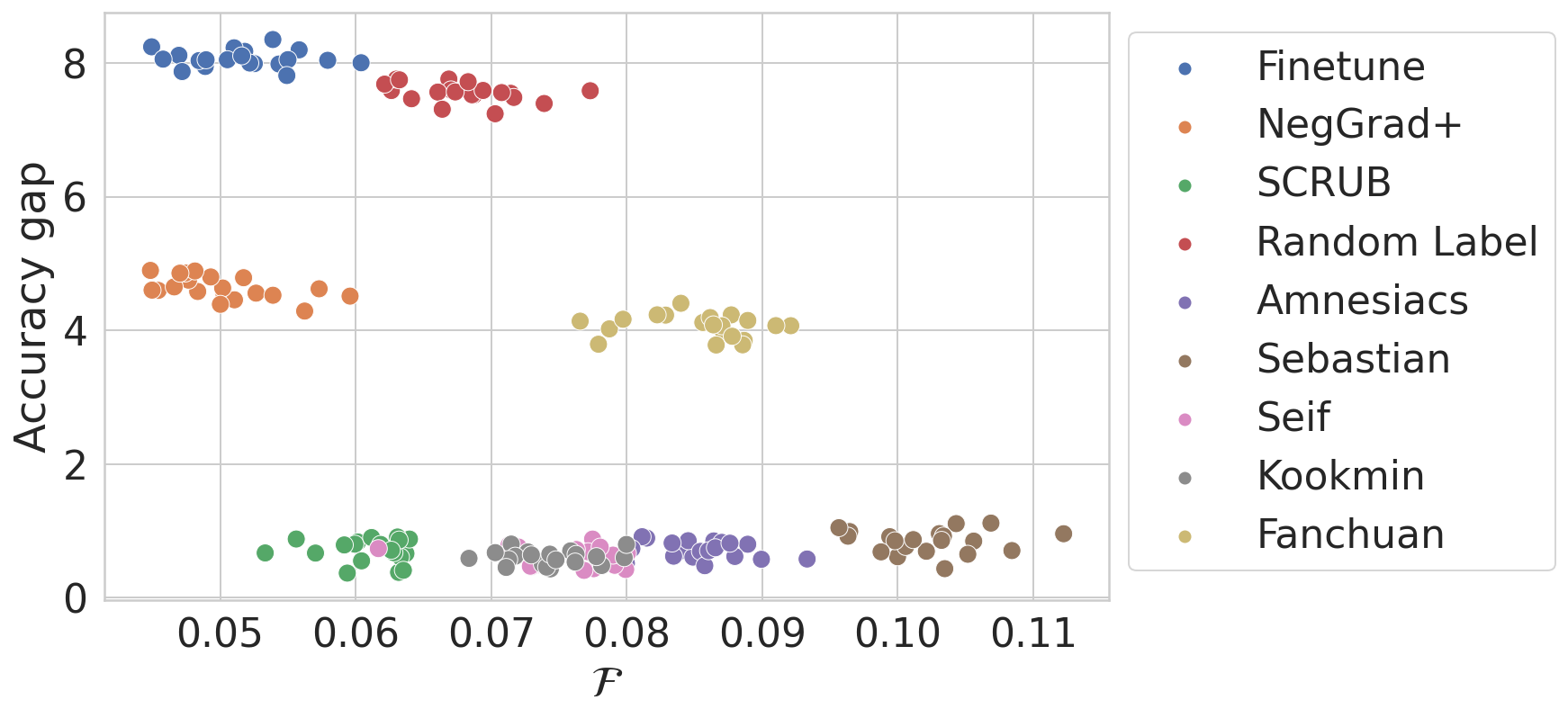}
\caption{The simple ``accuracy gap'' proxy (see Q4) for forgetting quality (smaller is better) versus our proposed $\mathcal{F}$-score (higher is better).}
\label{fig:correlation_with_forget_acc}
\end{wrapfigure}

\textbf{Q4. Relationship of $\mathcal{F}$-score with other metrics} Several metrics for forgetting quality have been proposed (see Section \ref{sec:related_work}). In Figure  \ref{fig:correlation_with_forget_acc}, we investigate the relationship between our $\mathcal{F}$-score and the simple ``accuracy gap'' metric, arguably the most commonly used in the literature. This measures the ``gap'' (absolute difference) between the (average) accuracy of the unlearned model and retrain-from-scratch, on the forget set (smaller is better). The Finetune baseline (with forget set accuracy 93.6\%) performs poorly in terms of both metrics: it has the lowest $\mathcal{F}$-score and the lowest ``gap'', due to overestimating the forget set accuracy of retrain (which is 85.6\%). Generally, the two metrics behave differently, with several methods that have similar ``accuracy gap'' having very different $\mathcal{F}$-scores; ``accuracy gap'' is not a good proxy for unlearning quality according to Definition \ref{defn:unlearning}.



\begin{figure}[b!]
\centering
\includegraphics[scale=0.3]{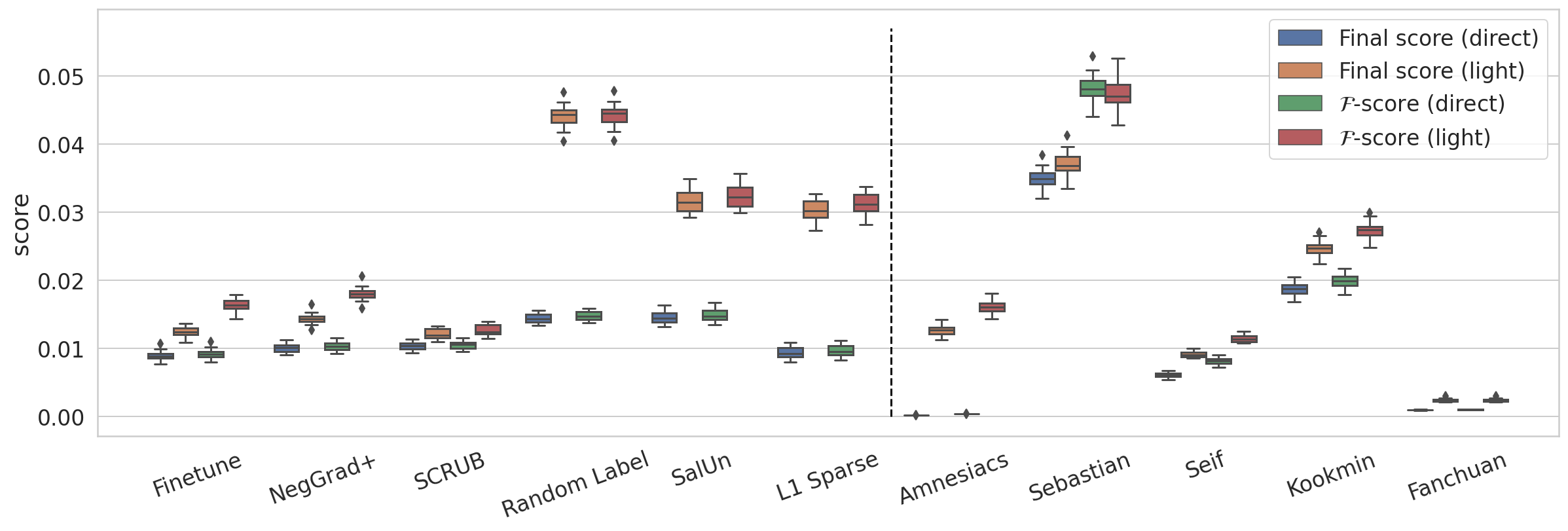}
\caption{The scores (see Equation \ref{eq:scores}) on FEMNIST when applying methods ``directly'' (with the same hyperparameters used for CASIA-SURF) or with light tuning (``light'').
}
\label{fig:femnist}
\end{figure}

\textbf{Q5. Generalizability} 
To investigate the ease of reusability of methods, we constructed an unlearning problem using a different dataset, Federated Extended MNIST (FEMNIST) \citep{caldas2018leaf}, which we modified to have similar properties and size as CASIA-SURF (see Section \ref{sec:femnist}), to maximize the chances of hyperparameter transfer. We inspect the performance of algorithms under two settings: i) ``directly'' using the hyperparameters tuned on CASIA-SURF, and ii) with ``light'' tuning around those values (via a grid with 3 values for each of 3 hyperparameters). We used setup ``Reuse-$N$-$N$'' here to save compute (Figure \ref{fig:setup_A_B_C} validates that it is a good proxy). Figure \ref{fig:femnist} reveals that some but not all top competition methods outperform existing ones under these minimal-adaptation settings. Out of existing methods, Random Label performs notably well under minimal tuning, outperforming some top competition entries. Sebastian remains one of the strongest methods (by far the best when applied directly out-of-the-box) perhaps due to its simplicity. We remark that poor performance under minimal adaptation on FEMNIST does not imply inability of a method to perform well on this dataset (if tuned extensively); and similarly, extensive tuning may lead to changes in the relative ranking of methods presented here. Instead, this experiment is designed to probe the generalizability and ease of reusability of methods, which we argue is a different but very important consideration.



\vspace{-2mm}
\section{Discussion and conclusion}
\vspace{-1mm}
In this paper, we presented a thorough evaluation of leading algorithms from the competition, and recent state-of-the-art methods under our proposed framework. We expanded on the initial instantiation of our framework that we used in the competition to consider variations that are more precise, at the expense of a higher compute cost. We found that several practical instantiations preserve the relative ranking of algorithms under the most compute intensive one, with important implications towards standardizing practical evaluation procedures.

However, there is important work remaining in designing accurate and efficient methods for evaluation. This is key: if applying an approximate method in practice requires evaluating it first in a manner that requires (much) more compute than retraining-from-scratch, this defeats the purpose of approximate unlearning in the first place. One may argue that we don't necessitate extensive evaluation for \textit{every} application of the same algorithm (e.g. on a new forget set), suggesting that we can amortize the cost. To achieve that, though, we would require strong indications that the performance on new forget sets or datasets does not change unexpectedly. We therefore emphasize generalizability of an algorithm (e.g. as exemplified by nearly-out-of-the-box adaptability on a new dataset) as an important property for unlearning and we identify strengths and weaknesses of different methods on this front.

So, are we making progress? We consider our benchmark a step forward for measuring the quality of an unlearning algorithm according to Definition \ref{defn:unlearning}, and we have shown that leading methods from the competition outperform existing ones according to our metric, showing substantial progress. 
However, we note that the competition methods were developed by iterating to improve our specific metric. In contrast, state-of-the-art methods may have been developed for different underlying applications for which other metrics are preferable or sufficient (e.g. we have seen that the ``accuracy gap'' metric correlates poorly with our proposed $\mathcal{F}$-score). 
Moving forward, we hope the community continues to identify key applications of unlearning and their underlying requirements in terms of defining and evaluating success, standardizing metrics where possible and continuing to build an understanding of algorithmic principles that are well-suited for ensuring strong performance on different metrics and subproblems. 

\newpage
\section{Acknowledgements} 
We thank Katja Filippova for her thoughtful feedback on the draft. We also acknowledge the competition teams of the top solutions \footnote{\url{https://www.kaggle.com/competitions/neurips-2023-machine-unlearning/leaderboard}}: fanchuan, [kookmin Univ] LD\&BGW\&KJH, Seif Eddine Achour, Sebastian Oleszko, toshi\_k \& marvelworld, Algorithmic Amnesiacs, Jiaxi Sun, Forget.


\bibliographystyle{plain}
\bibliography{neurips_data_2024}

\begin{thebibliography}{10}

\bibitem{attias2024information}
Idan Attias, Gintare~Karolina Dziugaite, Mahdi Haghifam, Roi Livni, and Daniel~M Roy.
\newblock Information complexity of stochastic convex optimization: Applications to generalization and memorization.
\newblock {\em arXiv preprint arXiv:2402.09327}, 2024.

\bibitem{bae2022if}
Juhan Bae, Nathan Ng, Alston Lo, Marzyeh Ghassemi, and Roger~B Grosse.
\newblock If influence functions are the answer, then what is the question?
\newblock {\em Advances in Neural Information Processing Systems}, 35:17953--17967, 2022.

\bibitem{barshan2020relatif}
Elnaz Barshan, Marc-Etienne Brunet, and Gintare~Karolina Dziugaite.
\newblock Relatif: Identifying explanatory training samples via relative influence.
\newblock In {\em International Conference on Artificial Intelligence and Statistics}, pages 1899--1909. PMLR, 2020.

\bibitem{bourtoule2021machine}
Lucas Bourtoule, Varun Chandrasekaran, Christopher~A Choquette-Choo, Hengrui Jia, Adelin Travers, Baiwu Zhang, David Lie, and Nicolas Papernot.
\newblock Machine unlearning.
\newblock In {\em 2021 IEEE Symposium on Security and Privacy (SP)}, pages 141--159. IEEE, 2021.

\bibitem{caldas2018leaf}
Sebastian Caldas, Sai Meher~Karthik Duddu, Peter Wu, Tian Li, Jakub Kone{\v{c}}n{\`y}, H~Brendan McMahan, Virginia Smith, and Ameet Talwalkar.
\newblock Leaf: A benchmark for federated settings.
\newblock {\em arXiv preprint arXiv:1812.01097}, 2018.

\bibitem{carlini2022membership}
Nicholas Carlini, Steve Chien, Milad Nasr, Shuang Song, Andreas Terzis, and Florian Tramer.
\newblock Membership inference attacks from first principles.
\newblock In {\em 2022 IEEE Symposium on Security and Privacy (SP)}, pages 1897--1914. IEEE, 2022.

\bibitem{cotogni2023duck}
Marco Cotogni, Jacopo Bonato, Luigi Sabetta, Francesco Pelosin, and Alessandro Nicolosi.
\newblock Duck: Distance-based unlearning via centroid kinematics.
\newblock {\em arXiv preprint arXiv:2312.02052}, 2023.

\bibitem{dwork2006differential}
Cynthia Dwork.
\newblock Differential privacy.
\newblock In {\em International colloquium on automata, languages, and programming}, pages 1--12. Springer, 2006.

\bibitem{fan2024challenging}
Chongyu Fan, Jiancheng Liu, Alfred Hero, and Sijia Liu.
\newblock Challenging forgets: Unveiling the worst-case forget sets in machine unlearning.
\newblock {\em arXiv preprint arXiv:2403.07362}, 2024.

\bibitem{fan2023salun}
Chongyu Fan, Jiancheng Liu, Yihua Zhang, Dennis Wei, Eric Wong, and Sijia Liu.
\newblock Salun: Empowering machine unlearning via gradient-based weight saliency in both image classification and generation.
\newblock {\em arXiv preprint arXiv:2310.12508}, 2023.

\bibitem{feldman2020does}
Vitaly Feldman.
\newblock Does learning require memorization? a short tale about a long tail.
\newblock In {\em Proceedings of the 52nd Annual ACM SIGACT Symposium on Theory of Computing}, pages 954--959, 2020.

\bibitem{goel2022towards}
Shashwat Goel, Ameya Prabhu, Amartya Sanyal, Ser-Nam Lim, Philip Torr, and Ponnurangam Kumaraguru.
\newblock Towards adversarial evaluations for inexact machine unlearning.
\newblock {\em arXiv preprint arXiv:2201.06640}, 2022.

\bibitem{goel2024corrective}
Shashwat Goel, Ameya Prabhu, Philip Torr, Ponnurangam Kumaraguru, and Amartya Sanyal.
\newblock Corrective machine unlearning.
\newblock {\em arXiv preprint arXiv:2402.14015}, 2024.

\bibitem{golatkar2020eternal}
Aditya Golatkar, Alessandro Achille, and Stefano Soatto.
\newblock Eternal sunshine of the spotless net: Selective forgetting in deep networks.
\newblock In {\em Proceedings of the IEEE/CVF Conference on Computer Vision and Pattern Recognition}, pages 9304--9312, 2020.

\bibitem{golatkar2020forgetting}
Aditya Golatkar, Alessandro Achille, and Stefano Soatto.
\newblock Forgetting outside the box: Scrubbing deep networks of information accessible from input-output observations.
\newblock In {\em Computer Vision--ECCV 2020: 16th European Conference, Glasgow, UK, August 23--28, 2020, Proceedings, Part XXIX 16}, pages 383--398. Springer, 2020.

\bibitem{gupta2021adaptive}
Varun Gupta, Christopher Jung, Seth Neel, Aaron Roth, Saeed Sharifi-Malvajerdi, and Chris Waites.
\newblock Adaptive machine unlearning.
\newblock {\em Advances in Neural Information Processing Systems}, 34:16319--16330, 2021.

\bibitem{hayes2024inexact}
Jamie Hayes, Ilia Shumailov, Eleni Triantafillou, Amr Khalifa, and Nicolas Papernot.
\newblock Inexact unlearning needs more careful evaluations to avoid a false sense of privacy.
\newblock {\em arXiv preprint arXiv:2403.01218}, 2024.

\bibitem{jagielski2020auditing}
Matthew Jagielski, Jonathan Ullman, and Alina Oprea.
\newblock Auditing differentially private machine learning: How private is private sgd?
\newblock {\em Advances in Neural Information Processing Systems}, 33:22205--22216, 2020.

\bibitem{kairouz2015composition}
Peter Kairouz, Sewoong Oh, and Pramod Viswanath.
\newblock The composition theorem for differential privacy.
\newblock In {\em International conference on machine learning}, pages 1376--1385. PMLR, 2015.

\bibitem{kodge2023deep}
Sangamesh Kodge, Gobinda Saha, and Kaushik Roy.
\newblock Deep unlearning: Fast and efficient training-free approach to controlled forgetting.
\newblock {\em arXiv preprint arXiv:2312.00761}, 2023.

\bibitem{koh2017understanding}
Pang~Wei Koh and Percy Liang.
\newblock Understanding black-box predictions via influence functions.
\newblock In {\em International conference on machine learning}, pages 1885--1894. PMLR, 2017.

\bibitem{kurmanji2024towards}
Meghdad Kurmanji, Peter Triantafillou, Jamie Hayes, and Eleni Triantafillou.
\newblock Towards unbounded machine unlearning.
\newblock {\em Advances in Neural Information Processing Systems}, 36, 2024.

\bibitem{liu2024model}
Jiancheng Liu, Parikshit Ram, Yuguang Yao, Gaowen Liu, Yang Liu, PRANAY SHARMA, Sijia Liu, et~al.
\newblock Model sparsity can simplify machine unlearning.
\newblock {\em Advances in Neural Information Processing Systems}, 36, 2024.

\bibitem{neel2021descent}
Seth Neel, Aaron Roth, and Saeed Sharifi-Malvajerdi.
\newblock Descent-to-delete: Gradient-based methods for machine unlearning.
\newblock In {\em Algorithmic Learning Theory}, pages 931--962. PMLR, 2021.

\bibitem{nguyen2022survey}
Thanh~Tam Nguyen, Thanh~Trung Huynh, Phi~Le Nguyen, Alan Wee-Chung Liew, Hongzhi Yin, and Quoc Viet~Hung Nguyen.
\newblock A survey of machine unlearning.
\newblock {\em arXiv preprint arXiv:2209.02299}, 2022.

\bibitem{paszke2019pytorch}
Adam Paszke, Sam Gross, Francisco Massa, Adam Lerer, James Bradbury, Gregory Chanan, Trevor Killeen, Zeming Lin, Natalia Gimelshein, Luca Antiga, et~al.
\newblock Pytorch: An imperative style, high-performance deep learning library.
\newblock {\em Advances in neural information processing systems}, 2019.

\bibitem{paul2021deep}
Mansheej Paul, Surya Ganguli, and Gintare~Karolina Dziugaite.
\newblock Deep learning on a data diet: Finding important examples early in training.
\newblock {\em Advances in Neural Information Processing Systems}, 34:20596--20607, 2021.

\bibitem{pawelczyk2023context}
Martin Pawelczyk, Seth Neel, and Himabindu Lakkaraju.
\newblock In-context unlearning: Language models as few shot unlearners.
\newblock {\em arXiv preprint arXiv:2310.07579}, 2023.

\bibitem{sablayrolles2019white}
Alexandre Sablayrolles, Matthijs Douze, Cordelia Schmid, Yann Ollivier, and Herv{\'e} J{\'e}gou.
\newblock White-box vs black-box: Bayes optimal strategies for membership inference.
\newblock In {\em International Conference on Machine Learning}, pages 5558--5567. PMLR, 2019.

\bibitem{sekhari2021remember}
Ayush Sekhari, Jayadev Acharya, Gautam Kamath, and Ananda~Theertha Suresh.
\newblock Remember what you want to forget: Algorithms for machine unlearning.
\newblock {\em Advances in Neural Information Processing Systems}, 34:18075--18086, 2021.

\bibitem{shah2023unlearning}
Vedant Shah, Frederik Tr{\"a}uble, Ashish Malik, Hugo Larochelle, Michael Mozer, Sanjeev Arora, Yoshua Bengio, and Anirudh Goyal.
\newblock Unlearning via sparse representations.
\newblock {\em arXiv preprint arXiv:2311.15268}, 2023.

\bibitem{shokri2017membership}
Reza Shokri, Marco Stronati, Congzheng Song, and Vitaly Shmatikov.
\newblock Membership inference attacks against machine learning models.
\newblock In {\em 2017 IEEE symposium on security and privacy (SP)}, pages 3--18. IEEE, 2017.

\bibitem{thudi2022unrolling}
Anvith Thudi, Gabriel Deza, Varun Chandrasekaran, and Nicolas Papernot.
\newblock Unrolling sgd: Understanding factors influencing machine unlearning.
\newblock In {\em 2022 IEEE 7th European Symposium on Security and Privacy (EuroS\&P)}, pages 303--319. IEEE, 2022.

\bibitem{wu2020deltagrad}
Yinjun Wu, Edgar Dobriban, and Susan Davidson.
\newblock Deltagrad: Rapid retraining of machine learning models.
\newblock In {\em International Conference on Machine Learning}, pages 10355--10366. PMLR, 2020.

\bibitem{zhang2020casia}
Shifeng Zhang, Ajian Liu, Jun Wan, Yanyan Liang, Guodong Guo, Sergio Escalera, Hugo~Jair Escalante, and Stan~Z Li.
\newblock Casia-surf: A large-scale multi-modal benchmark for face anti-spoofing.
\newblock {\em IEEE Transactions on Biometrics, Behavior, and Identity Science}, 2020.

\bibitem{zhaowhatmakes}
Kairan Zhao, Meghdad Kurmanji, George-Octavian Barbulescu, Eleni Triantafillou, and Peter Triantafillou.
\newblock What makes unlearning hard and what to do about it.
\newblock {\em arXiv preprint arXiv:2406.01257}, 2024.

\end{thebibliography}

\newpage

\appendix

\section{Appendix} \label{appendix}

\paragraph{Table of contents}
\begin{itemize}
    \item \ref{sec:limitations_and_impact} Discussion of limitations and negative societal impacts 
    \item \ref{sec:eval_in_detail}: Detailed description of our evaluation framework
    \item \ref{sec:details_competition}: Detailed description of the competition setup
    \item \ref{sec:detailed_algorithms}: Detailed description of unlearning algorithms
    \item \ref{sec:related_work}: Related work
    \item \ref{sec:femnist}: Federated Extended MNIST (FEMNIST)
    \item \ref{sec:implementation_details}: Implementation details
    \item \ref{sec:utility_forgetting_quality}: Utility / forgetting quality trade-off
    \item \ref{sec:per_example_breakdown}: Breaking down forgetting quality into per-example $\varepsilon$'s
    \item \ref{sec:stitch_erase_repair}: Stitching together different ``erase'' and ``repair'' phases
    \item \ref{sec:simple_mia}: Relationship between $\mathcal{F}$-scores and a simple MIA
    \item \ref{sec:histogram_visualizations}: Histograms of unlearned and retrained distributions
    \item \ref{sec:degree_of_overfitting}: Exploring the degree of overfitting the attacker 
    \item \ref{sec:effect_of_N}: Exploring the effect of $N$
    \item \ref{sec:bootstrapping}: Exploring bootstrapping to efficiently estimate $\mathcal{F}$
\end{itemize}

\subsection{Discussion of limitations and negative societal impacts} \label{sec:limitations_and_impact}

\paragraph{Limitations and future work} 
A limitation of our evaluation framework, and indeed any rigorous principled approach at evaluating unlearning, is its compute cost. As also discussed in the main paper, this can be important for practical application of unlearning algorithms, if frequent evaluation is required. We hope that the practical instantiations of our framework and the analyses we conducted pave the way towards better approaches that find a sweet spot in the spectrum of accuracy and compute cost.  

We have also discussed limitations of various lower-level design decisions of our framework in the main paper, and further in Section \ref{sec:eval_in_detail}. To summarize, some important future work directions on that front include investigating different design decisions for the attacks we carry out, the mechanism for aggregating estimated $\varepsilon$ values across examples as well as the mechanism for aggregating with utility and efficiency.

Further, due partially to the computationally intensive nature of our evaluation framework, we limit our investigation to two unlearning problems, one on the same dataset used for the competition, and another on FEMNIST. Similarly, we considered only one architecture and one training algorithm. We hope future work further explores different types of forget sets, datasets, architectures and training algorithms. However, we do feel that this work is an important step forward in benchmarking novel methods from the competition against previous state-of-the-art and has already surfaced previously-unknown findings both about evaluation as well as the strengths and weaknesses of new and existing algorithms.

Finally, we remark that, with unlearning being a young research area, the landscape is still forming regarding defining the problem and coming up with metrics for estimating success according to that definition. We don't claim that the definition we adopt, nor the evaluation framework we propose is well suited to every application of unlearning. For some problems one could perhaps use simpler proxies, like the ``Accuracy gap'' that we discussed in the main paper. We hope future work discusses this further and develops formal definitions and associated metrics for different subproblems under the umbrella of unlearning.

\paragraph{Societal impacts}
As with most technologies, unlearning can be used to both benefit and harm society.

Unlearning could be used to eliminate undesirable behavior (for example, by removing toxic examples), but it could also be used to remove ``good'' examples, yielding a model that is even more toxic or biased than before.

Additionally, unlearning could be used to eliminate defenses of large language models (LLMs). For example, if toxicity filters are learned by providing examples, then in theory, we could unlearn these to create a non-safe LLM.

We emphasize that this report is centered on the evaluation framework and examining the behaviors of existing and competitive algorithms. We did not identify any direct negative implications of this work.

\subsection{Detailed description of our evaluation framework} \label{sec:eval_in_detail}

In this section, we discuss our evaluation framework in greater detail and provide pseudocode that overviews the procedure (Algorithm \ref{alg:eval_overview}) and that details the per-example computation of $\varepsilon$ given estimated false positive and false negative rates of $m$ ``attacks'' (Algorithm \ref{alg:eps-from-fpr-fnr}). Our implementation is publicly available \footnote{\url{https://github.com/google-deepmind/unlearning_evaluation}}.

We note that several ``implementation details'' are important in computing $\varepsilon$ due to i) potential numerical issues and ii) obtaining robust results despite relatively few samples from each of the two distributions. We now discuss the choices we made. First, as shown in Algorithm \ref{alg:eps-from-fpr-fnr}, if both the FPR and FNR of an attack are equal to 0, we catch this as a special case and manually set the $\varepsilon$ to $\texttt{inf}$. This is because an attack that has both FPR and FNR equal to 0 perfectly separates the two distributions, indicating a complete failure for unlearning. On the other hand, if exactly one of FPR or FNR is 0 (but not both), we decided to discard this threshold, based on the assumption that this is an artifact of having insufficient samples from the two distributions. We hope that future work builds on our implementation and improves aspects of our framework.

Further, as mentioned in Section \ref{sec:forget_quality}, we used two families of attacks: single-threshold and double-threshold attacks. We found that including the latter family is really important, as we found in practice that unlearning algorithms produce different forms of distributions, not all of which are easily separable from retrain-from-scratch using only one threshold (see Section \ref{sec:histogram_visualizations} for some examples). One could of course further add more complex decision rules that use three or more thresholds, for instance. However, the potential downside there is that the increased complexity may lead to overfitting the particular samples rather than truly distinguishing the underlying distributions well.

We remark that we don't claim our set of attacks and their implementation is perfect, and we hope that future work improves on these. Our evaluation framework is general and agnostic to the particular choice of attacks, enabling easy plug-and-play. We hope that future work further experiments with different designs and analyzes the pros and cons of different instantiations from this perspective.

We also hope future work examines different aggregation strategies of per-example $\varepsilon$'s. We investigated different alternatives initially and considered e.g. returning the maximum $\varepsilon$ across examples, but we worried it would give too pessimistic an estimate, and we found it hard to distinguish different unlearning algorithms, because, while many are similar in the worst-case, their distributions of $\varepsilon$ values are different, and this would not be captured by this strategy. Computing quantiles over $\varepsilon$ values would also be possible, but we decided that the current proposal is a more granular way of comparing unlearning algorithms to one another. We hope that future iterations of our evaluation protocol will improve upon this choice.

\begin{algorithm}
\caption{Overview of evaluation for producing the final score for algorithm $\unlearn$ (w.r.t. $\alg$, $\forgetset$, $\dataset$)}\label{alg:eval_overview}
\begin{algorithmic}
\Require $\alg$, $\unlearn$, $\forgetset$, $\dataset$, $N$
\Require a procedure \texttt{compute-example-epsilon} that computes the $\varepsilon$ of an example by first applying a set of ``attacks'' aiming to distinguish the distribution of (processed, scalar) outputs of that example under unlearned vs retrained models, and then calling Algorithm \ref{alg:eps-from-fpr-fnr} to obtain $\varepsilon$ from the False Positive Rates and False Negative Rates of those attacks. 
\State $\theta^r_1 \dots \theta^r_N \sim \alg(\retainset)$ \Comment{Sample $N$ retrained models}
\State $\theta^o_1 \dots \theta^o_N \sim \alg(\dataset)$ \Comment{Sample $N$ original models}
\For{$i \in [1, N]$} \Comment{Turn each original model into an unlearned one}
    \State $\theta^u_i \gets \unlearn(\theta^o_i, \forgetset, \dataset)$
\EndFor
\State $\text{Retain-Acc}^r \gets \frac{1}{N} \sum_{i=1}^N\text{Acc}(\retainset, \theta^r_i), \ \ \ \text{Retain-Acc}^u \gets \frac{1}{N} \sum_{i=1}^N\text{Acc}(\retainset, \theta^u_i)$ 
\State $\text{Test-Acc}^r \gets \frac{1}{N} \sum_{i=1}^N\text{Acc}(\dataset_{test}, \theta^r_i), \ \ \ \ \ \ \ \text{Test-Acc}^u \gets \frac{1}{N} \sum_{i=1}^N\text{Acc}(\dataset_{test}, \theta^u_i)$

\State $\text{all-}\varepsilon \gets \{ \}$
\For{$s \in \forgetset$} \Comment{Compute per-example epsilons}
    \State $\varepsilon^s \gets \texttt{compute-example-epsilon}(s, \theta^r_1 \dots \theta^r_N, \theta^u_1 \dots \theta^u_N)$
    \State $\text{all-}\varepsilon\text{.add} (\varepsilon^s)$
\EndFor

\State $\mathcal{F} \gets \frac{1}{|\forgetset|} \sum_{\varepsilon \in \text{all-}\varepsilon}\mathcal{H}(\varepsilon)$  \Comment{Compute overall estimate of forgetting quality}

\State \Return $\mathcal{F} \times \displaystyle\frac{\text{Retain-Acc}^u}{\text{Retain-Acc}^r} \times \displaystyle\frac{\text{Test-Acc}^u}{\text{Test-Acc}^r}$  \Comment{Compute final score by adjusting for utility}

\end{algorithmic}
\end{algorithm}

\begin{algorithm}
\caption{Computes $\varepsilon^s$ for example $s \in \forgetset$ from the FPRs and FNRs obtained by carrying out $m$ attacks that aim to distinguish the unlearned and retrained distributions of (transformed) outputs for $s$. 
}\label{alg:eps-from-fpr-fnr}
\begin{algorithmic}
    \Require FPR, FNR: two lists of length $m$ each, storing the false positive and false negative rates (respectively) from running a collection of $m$ attacks to distinguish outputs for example $s$ of retrained and unlearned models. 
    \Require $\nanmax$: a function that returns the max of its inputs, discarding any that are nan. 
    \Require $\delta$: a float. 
    
    \State $\text{per-attack-}\varepsilon \gets \{ \}$
    
    \For{$i \in \{0 \dots m \}$}
    
    \If{$\text{FPR[i]} = 0$ and $\text{FNR[i]} = 0$} \Comment{Perfect separation of the two distributions.}
    \State $\text{per-attack-}\varepsilon \text{.add}(\texttt{inf})$
    
    \ElsIf{$\text{FPR[i]} = 0$ or $\text{FNR[i]} = 0$} \Comment{Discard this attack.}
    \State pass  
    
    \Else \Comment{Compute $\varepsilon$ via Equation \ref{eq:eps-from-fpr-fnr}}
    \State $\text{per-attack-}\varepsilon_1 \gets \log (1 - \delta - \text{FPR[i]}) - \log (\text{FNR[i])})$
    \State $\text{per-attack-}\varepsilon_2 \gets \log (1 - \delta - \text{FNR[i]}) - \log (\text{FPR[i])})$
    \State $\text{per-attack-}\varepsilon \text{.add}(\nanmax(\text{per-attack-}\varepsilon_1, \text{per-attack-}\varepsilon_2) ) $ 
    
    \EndIf
    
    \EndFor
    
    \State $\varepsilon \gets \nanmax (\text{per-attack-}\varepsilon)$ \Comment{The $\varepsilon$ for this example is that of the strongest attack}
    \State \Return $\varepsilon$
    
\end{algorithmic}
\end{algorithm}

\paragraph{Details of compute required by different instantiations}
For convenience, we present in Table \ref{tab:compute_of_instantiations} a breakdown of how compute-intensive each of our practical instantiations is, based on the number of samples that it requires from each of the original and retrained model distributions in order to obtain $E$ estimates of forgetting quality $\mathcal{F}$, each time using $N$ unlearned and $N$ retrained models (see Figure \ref{fig:setup_A_B_C}) for details.

\begin{table}
\begin{tabular}{ c c c }
    & Number of original models & Number of retrained models  \\
Setup ``Full'' & N $\times$ E &  N $\times$ E \\
Setup ``Reuse-$N$-$N$'' & N & N \\
Setup ``Reuse-$N$-1'' & 1 & N \\
Bootstrapping & K & K \\
\end{tabular}
\caption{Breakdown of how compute-intensive each of our practical instantiations is, based on the number of samples that it requires from each of the original and retrained model distributions in order to obtain $E$ estimates of forgetting quality $\mathcal{F}$, each time using $N$ unlearned and $N$ retrained models (see Figure \ref{fig:setup_A_B_C}) for details. For bootstrapping, we consider different values of the pool size $K$, but in all cases $K < E \times N$, and we use 8N by default, leading to significant savings over setup ``Full'' that requires E $\times$ N, where we set $E$ = 20 in our experiments.}
\label{tab:compute_of_instantiations}
\end{table}

\subsection{Detailed description of the competition setup} \label{sec:details_competition}

\paragraph{Overview} The competition was hosted on Kaggle\footnote{\url{https://www.kaggle.com/competitions/neurips-2023-machine-unlearning/}} and ran from September 11 to November 30, 2023. Submissions were automatically evaluated based on two key criteria: 1) the quality of forgetting (how effectively the model could remove specific information) and 2) the utility of the model (how well the model performed its intended task after unlearning), while we used a hard threshold on runtime to ensure efficiency.
This was a code-only competition, in which participants did not have access to the dataset nor the ``original'' and ``retrained-from-scratch'' models trained on that dataset. We also kept some details of our evaluation framework hidden during the competition: we did not reveal the attacks that we ran nor the form of function $h$ for processing the outputs. This decision was made in order to avoid participants developing approaches that were tailored to specific details of our evaluation procedure, rather than producing high quality unlearning algorithms that would perform well more generally, under different design decisions. Submissions were run on GPU equipped machines. For an entry to be considered valid, the algorithm had to execute the participant's unlearning algorithm across 512 model checkpoints within 8h. 

\begin{figure} 
    \centering  
    \includegraphics[width=0.8\linewidth]{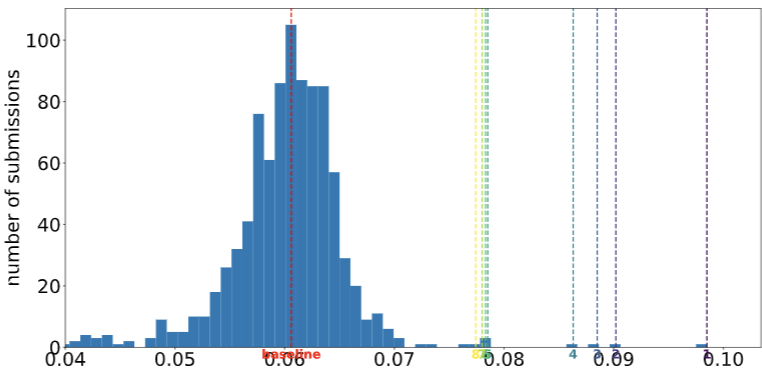} 
    \caption{Distribution of scores of submitted unlearning algorithms. Vertical lines denote the first 8 highest scoring submissions, with our baseline method in red.} 
    \label{fig:score_histogram} 
\end{figure}

\paragraph{Stats} A total of 1,338 individuals from 72 countries participated (i.e., made a submission to the leaderboard). For 500 of these participants (including 44 in the top 100), this was their first competition. At the end of the competition, there were 1,121 teams and 1,923 submissions. Figure \ref{fig:score_histogram} illustrates the distribution of final scores obtained by different submissions, with the red dotted line indicating the score of the ``Finetune'' baseline we provided in the starting notebook (which simply finetunes the original model on the retain set, relying on catastrophic forgetting to remove the influence of the forget set) \footnote{\url{https://www.kaggle.com/code/eleni30fillou/run-unlearn-finetune}}.

\paragraph{Evaluation setup for the competition} In more detail, the participants were asked to submit a Python notebook implementing their unlearning algorithm. The ``evaluation engine'' would then load the submitted algorithm, run it 512 times, starting each time from a single ``original model'' (Setup Reuse-$N$-1), and use those 512 unlearned models, together with (a single set of) 512 retrained models, in order to compute the score. In the terminology of this report, this corresponds to using Setup Reuse-$N$-1 with $N$ = 512 and $E$ = 1. We were forced to make these simplifications in the competition for efficiency and practicality reasons. In this report, we have additionally compared top methods (and state-of-the-art methods from the literature) under different instantiations of our evaluation framework that trade-off accuracy of the evaluation against computation cost.

\paragraph{Leaderboards} During the competition, the participants had access to a public leaderboard where they could see the score of their submissions. Participants developed algorithms to maximize this score. Then, to avoid overfitting on a particular retain/forget split, we finally re-evaluated each method on a version of the dataset splits created with a fresh random seed (the ``private split''). This private split was used to generate the final, ``private'', leaderboard that is now visible on kaggle and was used to award prizes. In this report, we analyze seven top methods according to the private leaderboard, using a fresh random seed that controls the retain / forget partition. 

\begin{figure}
     \centering
     \begin{subfigure}[b]{0.45\textwidth}
         \centering
         \includegraphics[width=\textwidth]{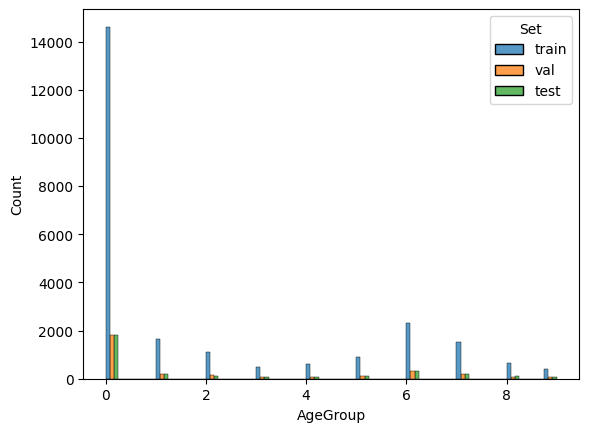}
         \caption{Train, validation and test histograms}
         \label{fig:train_val_test_hist}
     \end{subfigure}
     \hfill
     \begin{subfigure}[b]{0.45\textwidth}
         \centering
         \includegraphics[width=\textwidth]{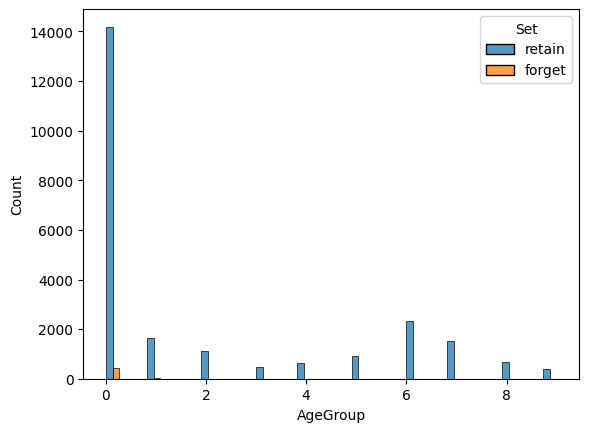}
         \caption{Retain and forget histograms}
         \label{fig:retain_forget_hist}
     \end{subfigure}
        \caption{Histograms of age groups for different sets. As can be seen, the distribution across classes / age groups is similar across the train, validation and test sets. There is a lot of class imbalance: class 0 is by far the most common. Note too that, while the retain set follows a similar distribution as the training set, the forget set contains examples from only the first two classes, with the vast majority belonging to class 0.}
        \label{fig:age_group_hist}
\end{figure}

\paragraph{Dataset details and forget set split} We use the CASIA-SURF dataset \citep{zhang2020casia} containing natural images of people's faces. Each image is labelled with an age group (there are 10 total classes / age groups). We split the dataset into a training, validation and test set. We further split the training set into a retain set and a forget set. When doing so, we take care that no subject's images are split between the retain and the forget set; that is, each subject is placed entirely in either the retain set or the forget set. The size of the forget set is roughly 2\% of the size of the training set. In Figure \ref{fig:age_group_hist}, we show the class (age group) distribution of the different sets.

\paragraph{Training details} The ``original model'' we consider is a ResNet-18 classifier, trained for 30 epochs on the training set to predict the age group associated with each image of a person's face. It is trained with class weights, to deal with class imbalance (where the loss value of an example is adjusted based on how frequent that example's class label is). We use no data augmentation. The original model obtains 98.98\% accuracy on the training set and 96.43\% on the test set.

\paragraph{Baseline unlearning algorithm} We consider a simple unlearning algorithm: finetuning the original model on only the retain set. We do this for 1 epoch using SGD with momentum of 0.9, a learning rate of 0.001 and weight decay of 5e-4.

\subsection{Detailed description of unlearning algorithms} \label{sec:detailed_algorithms}

\begin{figure}
    \centering
    \includegraphics[width=\textwidth]{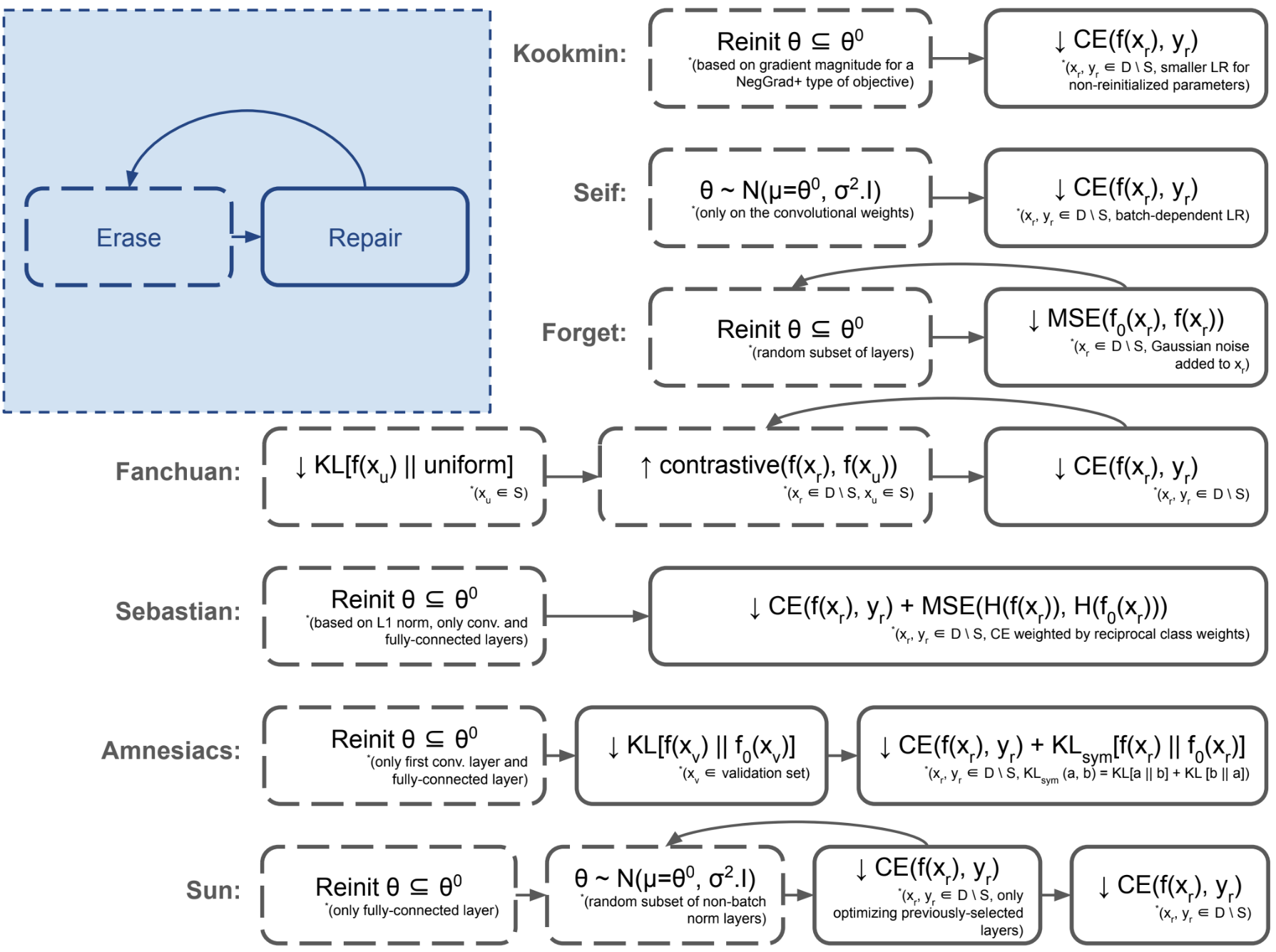}
    \caption{The top methods from the competition. ``Reinit'', ``CE'', ``MSE'', and ``H'' stand for ``reinitialize'', ``cross-entropy'', ``mean-squared error'', and ``entropy'', respectively. The upward ($\uparrow$) and downward ($\downarrow$) arrows indicate maximization or minimization of the objective, respectively.}
    \label{fig:methods_overview_full}
\end{figure}

As previously explained, the top methods from the competition operate in phases that can roughly be categorized into ``erasing'' the influence of the forget set and ``repairing'' any excess damage to the utility of the model that is caused by imperfect erasing (\autoref{fig:methods_overview_full}). We note that this is a separation that we find useful conceptually, though we remark that an operation can not be cleanly categorized into exactly one of ``erase'' or ``repair''. For instance, finetuning on the retain set can be seen as a means of repairing performance, but it also has an idirect effect of erasing information about the forget set passively, due to catastrophic forgetting. We categorize phases into ``erase'' and ``repair'' here based on their hypothesized \textit{primary} functionality.

The \href{https://www.kaggle.com/code/fanchuan/2nd-place-machine-unlearning-solution}{\bf Fanchuan} method 
first iterates over the entire forget set once and performs one step per mini-batch towards minimizing the KL-divergence between the predictions and a uniform distribution ($\downarrow\mathrm{KL}[f(x_u) \mid\mid \mathrm{uniform}]$). It then iterates for 8 epochs over the forget set and performs gradient ascent steps towards maximizing a (temperature-mitigated) dot-product contrastive loss between the forget set mini-batch and a mini-batch of randomly-sampled retain set examples ($\uparrow\mathrm{contrastive}(f(x_r), f(x_u))$). After each contrastive epoch, it performs one epoch of categorical cross-entropy training on the retain set ($\downarrow\mathrm{CE}(f(x_r), y_r)$).

The \href{https://www.kaggle.com/code/nuod8260/targeted-re-initialization/notebook}{\bf Kookmin} method  
reinitializes a subset of the model weights ($\mathrm{Reinit} \mkern9mu \theta \subseteq \theta^0$) before finetuning on the retain set ($\downarrow\mathrm{CE}(f(x_r), y_r)$). The parameters to be reinitialized are decided based on the gradient magnitude of the NegGrad+ loss over the forget and retain sets. The convolutional weights with the bottom 30\% gradient magnitudes are reinitialized. During finetuning, the gradients of the reinitialized and remaining convolutional parameters are multiplied by 1.0 and 0.1, respectively.

The \href{https://www.kaggle.com/code/seifachour12/unlearning-solution-4th-rank}{\bf Seif} method 
adds Gaussian noise ($\mu = 0, \sigma = 0.6$) to convolutional weights ($\theta \sim \mathcal{N}(\mu=\theta^0, \sigma^2 \cdot I)$) and performs 4 epochs of finetuning using a cross-entropy loss ($\downarrow\mathrm{CE}(f(x_r), y_r)$), the magnitude of which is adjusted based on the number of majority class examples present in the mini-batch. Rather than directly averaging the examplewise losses in the mini-batch, the {\bf Seif} method computes a weighted average of the examplewise losses using a weight of 1.0 for majority class examples and a weight of 0.05 for other examples. This is equivalent to using a learning rate which depends on the number of majority class examples in the mini-batch. Before the final epoch, additive Gaussian noise ($\mu = 0, \sigma = 0.005$) is applied to the convolutional weights.

The \href{https://www.kaggle.com/code/sebastianoleszko/prune-entropy-regularized-fine-tuning}{\bf Sebastian} method 
reinitializes a significant portion (99\%) of the convolutional and fully-connected layer weights with the lowest L1 norm ($\mathrm{Reinit} \mkern9mu \theta \subseteq \theta^0$), then performs finetuning on the retain set ($\downarrow\mathrm{CE}(f(x_r), y_r) + \mathrm{MSE}(H(f(x_r)), H(f_0(x_r)))$) using a combination of cross-entropy and mean squared error between the model prediction's entropy $H(f(x_r))$ and that of the original model $H(f_0(x_r))$.

The \href{https://www.kaggle.com/code/stathiskaripidis/unlearning-by-resetting-layers-7th-on-private-lb}{\bf Amnesiacs} method 
reinitializes the first convolutional layer and the fully-connected layer ($\mathrm{Reinit} \mkern9mu \theta \subseteq \theta^0$) before performing 3 ``warmup'' epochs of distilling the original model's predictions $f_0(x_v)$ for a held-out validation set into the reinitialized model ($\downarrow\mathrm{KL}[f(x_v) \mid\mid f_0(x_v)]$). The method then performs an additional 3 epochs of finetuning on the retain set using a combination of cross-entropy loss ($\mathrm{CE}(f(x_r), y_r)$) and symmetric KL-divergence loss ($\mathrm{KL}_\mathrm{sym}[f(x_r) \mid\mid f_0(x_r)]$) between the model's predictions $f(x_r)$ and the original model's predictions $f_0(x_r)$.

The \href{https://www.kaggle.com/code/sunkroos/noise-injection-unlearning-8th-place-solution}{\bf Sun} method 
reinitializes the fully-connected layer ($\mathrm{Reinit} \mkern9mu \theta \subseteq \theta^0$), then performs several epochs of ``noised'' finetuning on the retain set. Before each such epoch, a random subset of layers (excluding batch normalization) is selected and additive Gaussian noise is applied to their parameters ($\theta \sim \mathcal{N}(\mu=\theta^0, \sigma^2 \cdot I)$). The selected layers are then finetuned for an epoch. Finally, the model is finetuned normally on the retain set for a few epochs.

The \href{https://www.kaggle.com/code/jaesinahn/forget-set-free-approach-9th-on-private-lb}{\bf Forget} method 
iterates over several cycles of i) reinitializing a random subset of layers ($\mathrm{Reinit} \mkern9mu \theta \subseteq \theta^0$), and ii) distilling the original model's predictions on the forget set into the reinitialized model for an epoch using a mean squared error loss ($\downarrow\mathrm{MSE}(f_0(x_r), f(x_r)))$).

\subsection{Related work} \label{sec:related_work}
While unlearning is enjoying increased attention recently, it is a young area of research and the community lacks a standardized notion and associated evaluation metrics. In this work, we have adopted a a formal but non-worst case notion for unlearning that is largely inspired by \citep{sekhari2021remember, gupta2021adaptive, neel2021descent} and we have proposed an evaluation framework based on that. In this section, we describe other unlearning metrics that aren't necessarily tied to a formal definition, as well as different recent proposals for aggregating forgetting quality and utility.

\paragraph{Related metrics for forgetting quality} 
Several metrics for forgetting quality have been proposed based on the principle that unlearning should ideally ``match'' (different aspects of) a model retrained from scratch on only the retain set. While these share the same underlying goal, they vary substantially in complexity.

The simplest such metric that has been widely adopted in the context of classifiers is to simply report the ``accuracy gap'', i.e. the absolute difference of the accuracy of the unlearned model from the retrained model, on the forget set. Ideal unlearning according to this metric is characterized by an accuracy gap of 0. A different metric is based on ``relearn time'' \citep{golatkar2020eternal}: the ``time'' (in epochs) that it takes to relearn the forget set after having allegedly forgotten it. Intuitively, if the unlearned model can relearn the forget set much faster than a model that never trained on that set, this is an indication of imperfect unlearning according to this metric.

Instead, \cite{wu2020deltagrad} propose to measure forgetting quality via the $l_2$-distance in weight space between the unlearned and retrained models (a quantity that \cite{thudi2022unrolling} refer to this as the ``verification error''). \cite{thudi2022unrolling} further propose a proxy for the verification error, referred to as the ``unlearned error'' that is easy to compute and does not require the retrained-from-scratch model. Distances in weight space, however, are not very interpretable and may not be meaningful since neural networks are permutation-invariant and small variations to the training recipe (or even the order of mini-batches) may lead to different weights. \cite{golatkar2020eternal} instead use KL-divergence between the distributions of weights of the unlearned and retrained models, which accounts for the randomness of the training algorithm at the expense of being more computationally expensive. 

\cite{golatkar2020eternal} also discuss the ``Streisand effect'' where an unlearning method may cause the model's confidence on forget set examples to follow a very different distribution than would have been observed had the model never seen those examples. The Streisand effect refers to the undesired consequence of a sample becoming more noticeable after unlearning, causing vulnerability to attacks. These authors measure this effect qualitatively by plotting the distribution of the entropy of the unlearned model outputs. Ideally, this distribution would match that of the retrained-from-scratch model.

To operationalize this intuition, several Membership Inference Attacks (MIAs) of varying degrees of complexity have also been proposed: if an attacker can infer that a sample has been unlearned (rather than not having ever been trained on), this marks a failure for the unlearning algorithm. Attackers of varying degrees of strength have been used in the unlearning literature. The MIAs proposed in \citep{golatkar2020eternal,golatkar2020forgetting,liu2024model,kodge2023deep} and the ``Baseline MIA'' in \citep{kurmanji2024towards} are simple attacks instantiated as binary classifiers (e.g. using logistic regression) that operate directly on outputs (e.g. entropies or confidences) of the unlearned model. For instance, the attack in \citep{golatkar2020eternal} trains a binary classifier to separate outputs from the retain set (``in'') versus the test set (``out'') and then queries this classifier on outputs of the forget set (ideally predicted as being ``out''). On the other hand, \citep{kurmanji2024towards}'s ``Baseline MIA'' trains a binary classifier that directly aims to distinguish outputs of the forget and test sets. Ideally, this classifier (applied on ``held-out'' samples from the forget and test sets) should struggle to separate the two sets and shouldn't perform any better than if it were applied on a model retrained from scratch. 

Recently, unlearning papers have also started using stronger MIAs, inspired by LiRA \citep{carlini2022membership}, a state-of-the-art MIA in the privacy community. This MIA can be thought of as instantiating a dedicated attacker for each example, that infers the membership status of that example (``in'' versus ``out'' in the classical version; ``unlearned'' versus ``out'' in the adaptation to unlearning) based on information tailored to that specific example. In particular, through training several shadow models, LiRA collects confidences of each particular example under the two different ``worlds'' and uses those to estimate one Gaussian for each of the two worlds for a given example, allowing to make predictions based on likelihood under those Gaussians. \cite{kurmanji2024towards,pawelczyk2023context} recently reported results with an adaptation of LiRA for unlearning, and \cite{hayes2024inexact} demonstrated that this attack is substantially stronger than previous ones used in the literature, that unfortunately had overestimated the privacy protection afforded by unlearning algorithms. This attack is the closest to our evaluation framework. 

\paragraph{Alternative notions and metrics} As a significant departure from the spirit of the above notions and metrics, some authors argue that, for some applications, matching retraining-from-scratch is not necessarily the criterion of interest. Instead, unlearning may be employed to mitigate the impact of incorrect of adversarially-manipulated training data, a subset of which may be known (this is referred to as ``corrective unlearning'', in \citep{goel2024corrective}). Dedicated metrics can be defined for these applications that capture the reversal of unwanted behaviours that were learned from the problematic data. \cite{goel2022towards}, for instance, propose a metric that measures the degree of ``confusion'' between two classes that is due to the presence of mislabelled examples between those classes in the original training set. A successful unlearning algorithm, operating on a forget set containing all and only the mislabelled training data, would fully eliminate that confusion. \cite{shah2023unlearning} similarly aim for the goal of what they refer to as ``complete unlearning'': achieving minimal accuracy on the forget set (without harming utility), irrespective of the accuracy of retraining-from-scratch. \cite{kurmanji2024towards} argues that the notion and metric of unlearning should be application-dependent e.g. unlearning old data simply for keeping the model up to date may come with vastly different desiderata or priorities than unlearning to protect user privacy. They propose a method that can be adapted to handle different scenarios and investigate empirically different metrics for different applications.

\paragraph{Alternative aggregation strategies} Regardless of how one defines and measures ``forgetting quality'', a holistic evaluation of unlearning requires taking into account utility and efficiency as well. While in most research papers these aspects have been explored in isolation, some recent works propose (partial) aggregation mechanisms, for utility and forgetting quality. 

Specifically, \cite{fan2023salun} propose the ``Average Gap'' metric. In particular, for a given metric (e.g. the accuracy on the forget set), they compute the ``gap'' (absolute difference) between the unlearned and retrained models' performance according to that metric. Then, they average the ``gaps'' across metrics that capture both forgetting quality as well as utility. In their work, they average the gaps for accuracy-based metrics only, where the gap in terms of forget set accuracy captures forgetting quality and the gaps in terms of retain and test accuracy capture utility. \cite{cotogni2023duck} also propose an aggregate metric called Adaptive
Unlearning Score (AUS), that combines an estimate of forgetting quality (in their work this is given by the accuracy on the forget set) with the performance loss relative to the original model, measured in terms of test accuracy, as an estimate of utility.  \cite{zhaowhatmakes} propose a ``Tug of War'' score, where higher is better, obtained by also combining ``average gaps'' on the forget, retain and test sets, aiming to capture trade-offs.

Generally, the unlearning literature lacks discussion about aggregation strategies and it not clear a priori when one of the aforementioned proposals is preferable over another. Designing careful aggregation strategies that allow us to both assess unlearning quality holistically and encode trade-offs in an interpretable manner is an interesting study for future work.

\subsection{Federated Extended MNIST (FEMNIST)} \label{sec:femnist}

We chose the Federated Extended MNIST (FEMNIST) dataset \cite{caldas2018leaf} as the second dataset we use in this report due to its similar structure to CASIA-SURF. Specifically, this dataset contains a set of ``users'', each of which has hand-drawn a set of number and characters. The classification problem is a 62-way task, to distinguish between numbers and characters written by different users, based on 28x28 color images. We adjusted the dataset by (randomly) picking a subset of size 30400. This is in order to have the same dataset size as CASIA-SURF, to increase the chance of hyperparameters tranferring. This is key due to our experimental setup and underlying research question that we investigate on FEMNIST, namely probing at the generalizability and ease of reusability of different unlearning algorithms.

We followed the same protocol in creating different splits in FEMNIST as we did in the competition for CASIA-SURF, again to ensure as similar properties as possible between the two. Specifically, we created an 80\% / 10\% / 10\% train / validation / test split uniformly at random. Then, we divided the training set into a retain / forget partition by ensuring that there is no overlap in users between the retain and forget sets, reflecting a scenario where a subset of users request their data to be deleted. The sizes of the retain and forget set are 23853 and 467, respectively.

\begin{figure}
    \centering
    
    \begin{subfigure}[b]{0.3\textwidth}
         \includegraphics[scale=0.27]{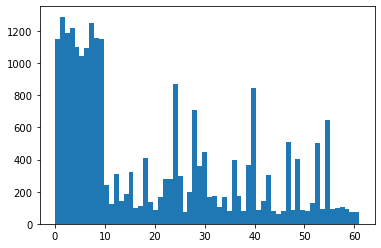}
         \caption{Retain set.}
         \label{fig:femnist_retain}
    \end{subfigure}
    \begin{subfigure}[b]{0.3\textwidth}
         \includegraphics[scale=0.27]{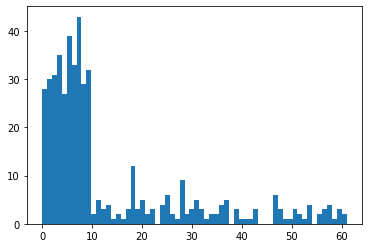}
         \caption{Forget set.}
         \label{fig:femnist_forget}
    \end{subfigure}
    \begin{subfigure}[b]{0.3\textwidth}
         \includegraphics[scale=0.27]{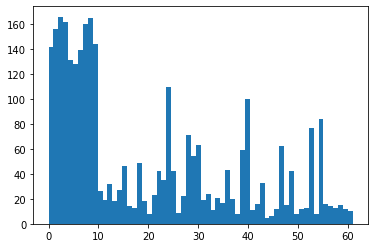}
         \caption{Test set.}
         \label{fig:femnist_test}
    \end{subfigure}
    
    \caption{Label distribution of FEMNIST.}
    \label{fig:label_distributions_femnist}
\end{figure}

Similar to CASIA-SURF, this dataset also has class imbalance. We plot the label distribution of the retain, forget and test sets in Figure \ref{fig:label_distributions_femnist}.

Refer to the below section for implementation details on both CASIA-SURF and FEMNIST.

\subsection{Implementation details} \label{sec:implementation_details}

We implemented all experimentation in this report by further building on top of our public code-base \footnote{\url{https://github.com/google-deepmind/unlearning\_evaluation}} (released under the Apache 2.0 license), in PyTorch \citep{paszke2019pytorch}.

For the state-of-the-art baselines, we adapted publicly-available code from the respective authors, to ensure correctness, largely using the repository associated with SalUn \cite{fan2023salun} \footnote{\url{https://github.com/OPTML-Group/Unlearn-Saliency}}, which also implements other baselines. For the Random Label baseline, we follow the implementation that operates in a sequence, first finetuning on the forget set with the random labels, and subsequently finetunes on the retain set (a ``repair phase'', similar in spirit to top submissions).

\subsubsection{Hyperparameters for original training} For all experiments, we used the same architecture as in the competition, a ResNet-18. For our CASIA-SURF experiments, we follow the same training details as for the competition, as described in Section \ref{sec:details_competition}. We trained for 30 epochs, using SGD with momentum 0.9, weight decay of 5e-3 and a learning rate of 0.0001.
For FEMNIST, we found we needed many more epochs to obtain strong performance, perhaps due to the much larger set of classes involved in this task. We trained for 120 epochs, using SGD with momentum 0.9, a weight decay of 0.001 and a learning rate of 0.0005.

For both datasets, to deal with the imbalance, we calculated a ``weight'' for each class as the reciprocal number of occurrences of that class in the training set, so that popular classes are associated with a smaller such weight compared to rare classes. We then passed the class weights to pytorch's ``CrossEntropyLoss'' function as the ``weight'' parameter, leading to down-weighting popular classes during training, in order to enable learning the rare classes too. The original model on SURF obtains 98.98\% training accuracy and 96.43\% test accuracy, and the original model on FEMNIST obtains 98.95\% training accuracy and 95.76\% test accuracy.

\subsubsection{Hyperparameters for unlearning} 
\paragraph{CASIA-SURF} For all analyses in the CASIA-SURF dataset, we used the competition algorithms using the configurations from the respective submissions. 

On the other hand, comparing against state-of-the-art methods necessitates tuning those methods on CASIA-SURF, for a fair comparison. We remark that competition methods have a strong advantage in this comparison: they were developed by iterating to improve a score that reflects the exact metric used for evaluation. To give the state-of-the-art methods a fair chance, we tuned their hyperparameters ``optimistically'', by computing the final score (see Equation \ref{eq:scores}) of different hyperparameter settings, and choosing the setting with the highest score. To make this practical, we used Setup ``Reuse-N-N'' for hyperparameter tuning, with $N = 512$. We repeated this $10$ times for each configuration and picked the configuration that had the best average final score. 

We note that this tuning procedure is not always practical given that computing our score is expensive. We chose to use it here for a fair reflection of whether competition methods improve upon the state-of-the-art; having tuned state-of-the-art methods less optimistically may mislead us in incorrectly answering the above question to the affirmative. Nonetheless, we hope that future work iterates on approaches for model selection and hyperparameter tuning that are computationally efficient, leading to practical unlearning pipelines.

More concretely, for each method considered, we utilized a grid of 3 values for each of 3 hyperparameters. We list the considered values below:
\begin{itemize}
    \item NegGrad+: We tuned the learning rate (0.0005, 0.001, 0.005), the number of epochs (1, 2, 3) and the $\alpha$ parameter that balances the retain and forget losses (0.999 0.9999 0.99999); see \citep{kurmanji2024towards} for details. The best values we discovered were 0.001, 1 and 0.99999, respectively.
    
    \item SCRUB: We tuned the learning rate (0.0001, 0.0005, 0.001), the number of epochs (1, 2, 3) and the number of ``max steps'' (1, 2, 3), i.e. epochs in which to perform both ``min'' and ``max'' steps rather than just ``min''. Note that any trailing ``min'' steps can be seen as a ``repair'' phase, in line with the sequential nature of several competition methods; see \citep{kurmanji2024towards} for details. The best values we discovered were 0.0005, 1 and 1, respectively.
    
    \item Random Label: We tuned the learning rate of the erase phase (0.0001, 0.0005, 0.001), the number of epochs of the erase phase (1, 2, 3) and for the repair phase (1, 2, 3). The best values we discovered were 0.0005, 2, and 1, respectively.
    
    \item SalUn: We tuned the threshold for choosing which parameters will be masked based on their gradient magnitude (0.4, 0.5, 0.6), the learning rate (0.0001, 0.0005, 0.001) and number of epochs (1, 2, 3) of the erase phase of the Random Label algorithm on which SalUn is based; see \citep{fan2023salun} for details. The best values we discovered were 0.5, 0.0005 and 2, respectively. For other Random Label hyperparameters, we used the best values found for Random Label.
    
    \item L1 Sparse: We tuned the learning rate (0.0005, 0.001, 0.005) and $\alpha$ parameter controlling the weight of the L1-penalty (1e-5, 1e-4, 1e-3, 1e-2, 1e-1); see \citep{liu2024model} for details. Given there were only two relevant hyperparameters in this case, we allowed more values for the second. The best values discovered were 0.001 and 0.0001, respectively.
\end{itemize}


\paragraph{FEMNIST} For the experiments on FEMNIST, we tuned hyperparameters both for the competition methods and state-of-the-art methods. In all cases, we tuned around the values that worked best for CASIA-SURF, using again a grid of 3 values for each of 3 hyperparameters per method.

We note that the cross-entropy reweighting strategy that Seif uses is hard-coded for 10 classes (and based on the label distribution of CASIA-SURF) and is therefore not applicable to FEMNIST. We applied it on FEMNIST with a vanilla cross-entropy, without reweighting, recognizing that the performance of Seif on FEMNIST could be improved by other strategies.

We list the considered values below:
\begin{itemize}
    \item Fanchuan: We tuned the learning rate of the first erase phase (0.001, 0.005, 0.01), the learning rate of the second erase phase (1e-4, 3e-4, 6e-4) and the learning rate of the repair phase (0.0005, 0.001, 0.005). We were not able to discover good hyperparameter values within this grid.
    \item Kookmin: We tuned the learning rate of the initial phase used to determine which parameters to reinitialize (0.1, 0.3, 0.5), the learning rate of the repair phase (0.0005, 0.001, 0.005) and number of epochs (4, 5, 6).The best values discovered were 0.3, 0.0005 and 5, respectively.
    \item Seif: We tuned the standard deviation of the Gaussian for adding noise (0.5, 0.6, 0.7), the learning rate (0.0004, 0.0007, 0.001) and number of epochs (3, 4, 5). The best values discovered were 0.6, 0.001 and 5, respectively.
    \item Sebastian: We tuned the learning rate (0.0001, 0.0005, 0.001), the pruning amount (0.999, 0.99, 0.9) and the epochs (2.2, 3.2, 4.2), represented as a float in their implementation. The best values discovered were 0.0005, 0.99 and 4.2, respectively.
    \item Amnesiacs: We tuned the learning rate for the warmup phase (8e-4, 9e-4, 1e-3), the learning rate for the repair phase (5e-4, 1e-3, 5e-3) and number of epochs (2, 3, 4). The best values discovered were 0.001, 0.005 and 4, respectively.
\end{itemize}

The best values we discovered for the state-of-the-art algorithms on this dataset were the following:
\begin{itemize}
    \item NegGrad+: learning rate of 0.005, 1 epoch and $\alpha$ 0.9999.
    \item SCRUB: learning rate of 0.001, 2 epochs and 2 max steps.
    \item Random label: learning rate for the forgetting phase 0.001, 2 epochs for each of the erase and repair phases.
    \item SalUn: threshold of 0.4, learning rate 0.001 and 2 epochs.
    \item L1 Sparse: learning rate 0.001 and $\alpha$ 0.001.
\end{itemize}

\subsection{Utility / forgetting quality trade-off} \label{sec:utility_forgetting_quality} 
\begin{figure}[ht]
\centering
\includegraphics[scale=0.25]{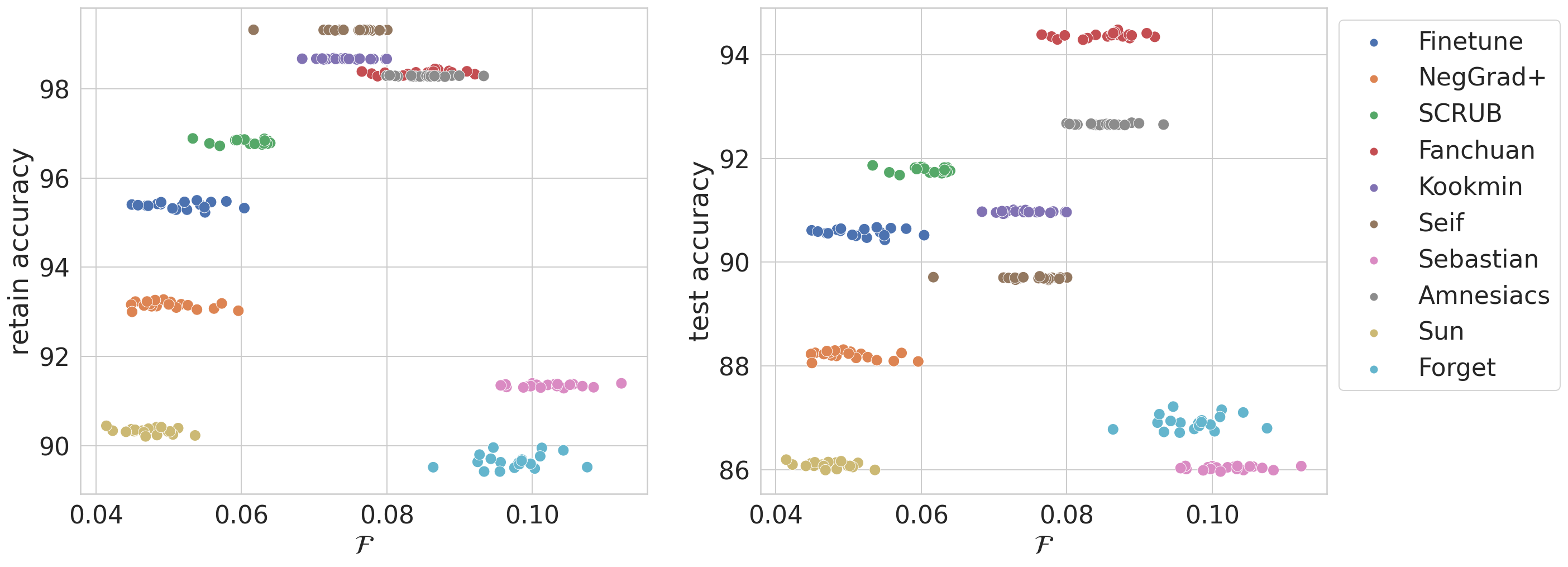}
\caption{The utility / forgetting quality trade-off where utility is measured in terms of the retain and test accuracy. 
We notice that some unlearning methods harm retain or test accuracy disproportionately more than the other (e.g. Kookmin and Seif have high retain but low test accuracy while Forget has lower retain than test accuracy). We also observe diverse trade-off profiles, with some methods (like SCRUB) having high retain and test accuracy but low $\mathcal{F}$-score, while, as discussed previously, Sebastian has the highest $\mathcal{F}$-score but poor utility. We observe that while there is some variance in the $\mathcal{F}$-score across different runs of a given unlearning algorithm, the retain and test accuracy of these runs have smaller variance.
}
\label{fig:utility-forgetting-tradeoff}
\end{figure}

We further investigate the trade-off between utility and forgetting quality in Figure \ref{fig:utility-forgetting-tradeoff}. We notice that some unlearning methods harm retain or test accuracy disproportionately more than the other (e.g. Kookmin and Seif have high retain but low test accuracy while Forget has lower retain than test accuracy). We also observe diverse trade-off profiles, with some methods (like SCRUB) having high retain and test accuracy but low $\mathcal{F}$-score, while, as discussed previously, Sebastian has the highest $\mathcal{F}$-score but poor utility.

\subsection{Breaking down forgetting quality into per-example \texorpdfstring{$\varepsilon$}{eps}'s} \label{sec:per_example_breakdown} We now inspect the distribution of per-example $\varepsilon$'s produced by different algorithms. We are interested in investigating: are better performing algorithms better due to uniformly improving $\varepsilon$'s of all examples equally, or do they improve by further boosting a subset of the $\varepsilon$'s by a larger amount? Are there examples that are difficult to unlearn, across all algorithms?

\begin{figure}[ht]
    \centering
    \includegraphics[scale=0.3]{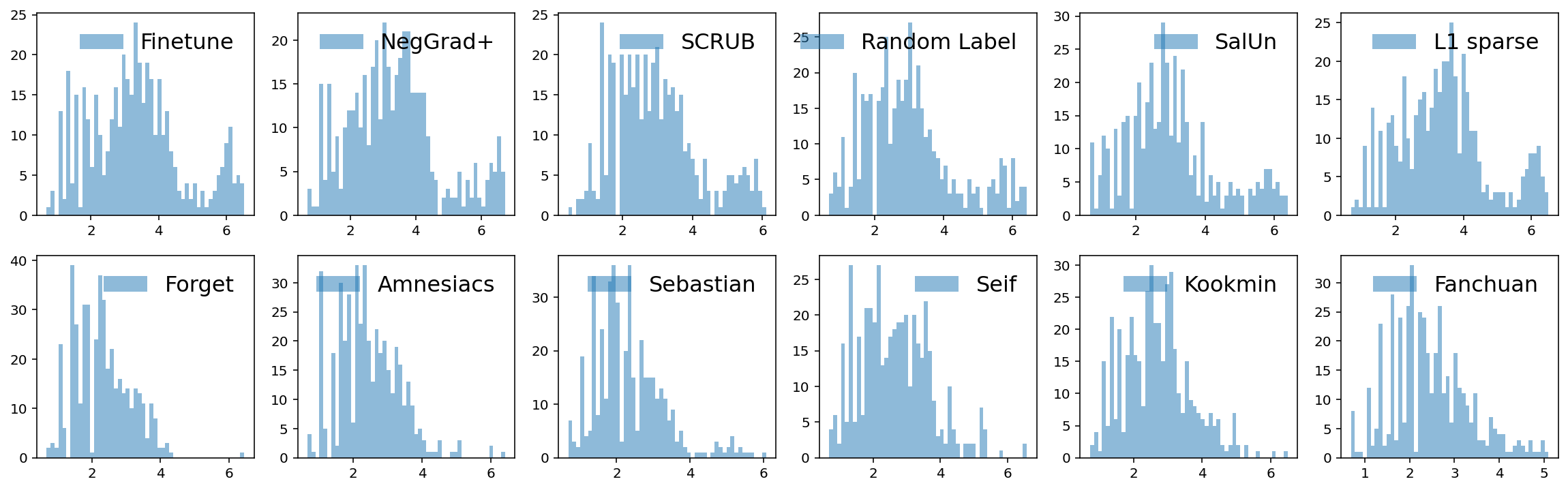}
    \caption{For different unlearning algorithms, the histogram of per-example $\varepsilon$.}
    \label{fig:eps_hists}
\end{figure}

In Figure \ref{fig:eps_hists}, we visualize histograms of the distribution of per-example $\varepsilon$'s, for different unlearning algorithms. We observe that several algorithms span a wide range of $\varepsilon$ values, but we do see differences between their distributions. For instance, the Finetune baseline leads to several examples having $\varepsilon$ values near the higher end, which is not the case for better-performing algorithms. Interestingly, Forget (which is one of the top methods in terms of forgetting quality $\mathcal{F}$) has very few $\varepsilon$ values greater than 4, which is not the case for other top methods. We believe these characteristics warrant further investigation and understand better properties of different algorithms is very valuable, to inform which to choose for different downstream applications (e.g. where we care about worst-case versus average-case performance).

\begin{figure}[ht]
    \centering
    \includegraphics[scale=0.3]{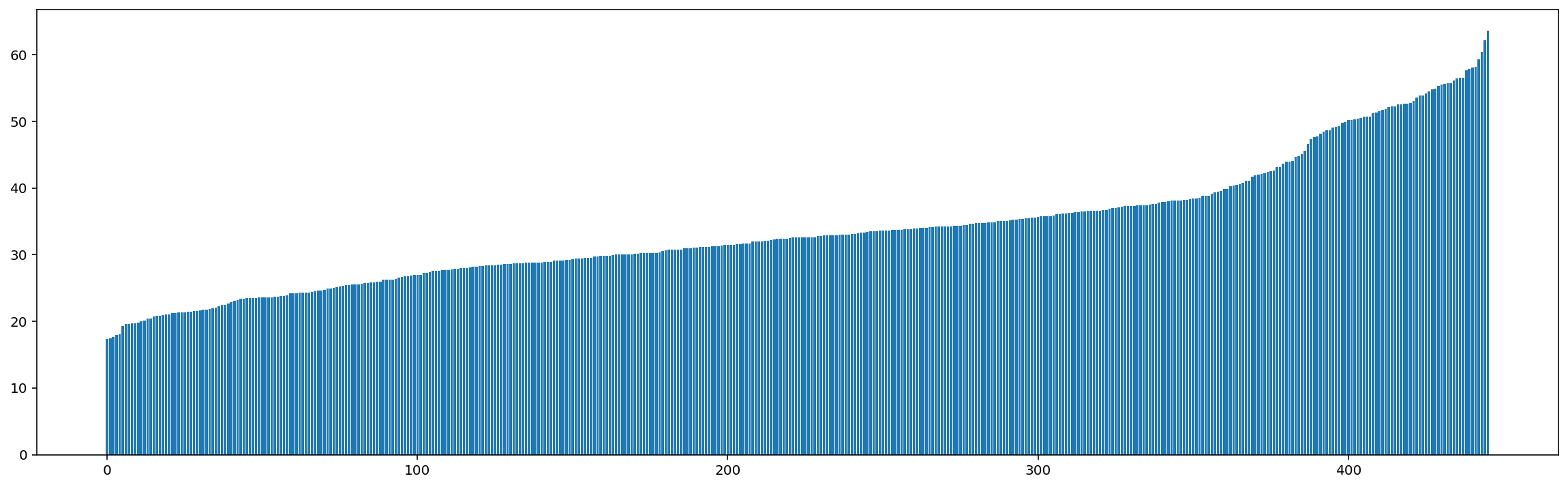}
    \caption{For each example in the forget set (represented by a different bar in this barplot), the sum of the $\varepsilon$ values for that example across the twelve unlearning methods used in \ref{fig:eps_hists}. We sort examples by their sum for easier visualization. We observe (at the far right), that there are some examples approaching the maximum realizable value of this sum (which is roughly 6 x 12 = 72; since there are 12 methods and the maximum $\varepsilon$ value is roughly 6). This indicates that these examples are hard for all algorithms considered. On the other hand, the far left of the plot reveals the existence of some overall easy examples too.}
    \label{fig:example_difficulty}
\end{figure}

We further conduct a preliminary investigation on whether there are certain examples that are ``hard'' (i.e. associated with a large $\varepsilon$) across unlearning algorithms. In Figure \ref{fig:example_difficulty} we show, for each example in the forget set (represented by a different bar in this barplot), the sum of the $\varepsilon$ values for that example across the twelve unlearning methods used in \ref{fig:eps_hists}. We sort examples by their sum for easier visualization. We observe (at the far right), that there are some examples approaching the maximum realizable value of this sum (which is roughly 6 x 12 = 72; since there are 12 methods and the maximum $\varepsilon$ value is roughly 6). This indicates that these examples are hard for all algorithms considered. On the other hand, the far left of the plot reveals the existence of some overall easy examples too.

Recent work \cite{fan2024challenging,zhaowhatmakes} discusses the creation of challenging forget sets, by treating the problem as adversarial optimization, and investigating hypotheses surrounding interpretable factors that influence difficulty, respectively. We leave it to future work to apply those methods to the novel algorithms developed in the competition, and under our evaluation framework.

\begin{figure}[ht]
    \centering
    \includegraphics[scale=0.28]{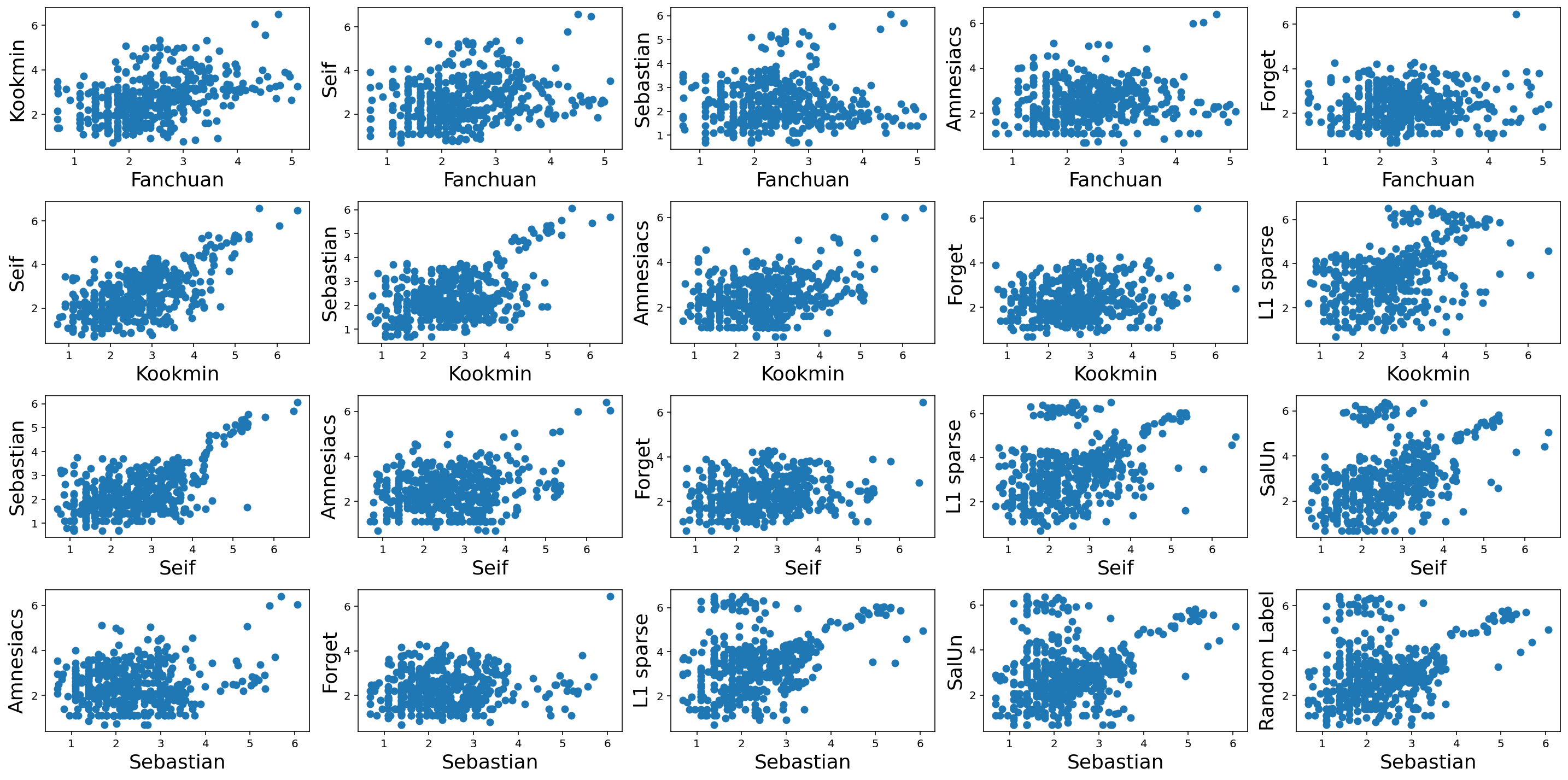}
    \caption{Relationship between $\varepsilon$ values produced by different pairs of algorithms. Each subplot corresponds to a pair of algorithms and contains one dot per example in the forget set, where the $x$-value is the $\epsilon$ for that example for one algorithm and the $y$-value is the $\epsilon$ for that same example for the other algorithm.}
    \label{fig:eps_pairwise}
\end{figure}

Further, we look into correlations of $\varepsilon$ values between pairs of algorithms. We plot the results in from Figure \ref{fig:eps_pairwise}. Each subplot corresponds to a pair of algorithms and contains one dot per example in the forget set, where the $x$-value is the $\epsilon$ for that example for one algorithm and the $y$-value is the $\epsilon$ for that same example for the other algorithm. We notice that different pairs of algorithms have different strengths of correlation between their $\varepsilon$ values estimated for different examples. We do see, for instance, that Sebastian and Kookmin agree on some hard examples, but we also notice several disagreements. For instance, L1 Sparse, SalUn and Random Label each have a cluster of examples with high $\varepsilon$ that have lower $\varepsilon$ values according to Sebastian. An interesting observation is also that Sebastian and Forget (second cell of the bottom row), the two top methods in terms of $\mathcal{F}$-scores, don't agree much on which examples are the hardest. We hope future work investigates these phenomena further.

\subsection{Stitching together different ``erase'' and ``repair'' phases} \label{sec:stitch_erase_repair} In this section, we investigate whether one can directly ``stitch'' the erase phase of one unlearning algorithm together with the ``repair'' phase of another, or whether the erase and repair phases have co-adapted (e.g. their hyperparameters and design principles depend heavily on one another).

Specifically, we look into whether we can directly improve the performance of unlearning algorithms that have a ``vanilla'' repair phase (simply finetuning on the retian set with a standard cross-entropy loss), by replacing their repair phase with a more sophisticated repair phase from another unlearning algorithm. We seek to do this in the simplest form, without updating any hyperparameters, as a proof of concept.

To that end, we take two unlearning algorithms that have a vanilla repair phase, Fanchuan and Kookmin, and we investigate the effect of replacing that repair phase with Seif's, that can be seen as applying a per-batch adjustment to the learning rate based on how many examples of the majority class are in the current batch (batches with many majority class examples will receive a \textit{larger} effective learning rate compared to batches with fewer majority class examples); see Section \ref{sec:detailed_algorithms} for a more detailed explanation.

When integrating Seif's repair phase into Fanchuan and Kookmin, we adjust their (base) learning rate for the repair phase so that, on average, the per-batch effective learning rates will be the same as the original learning rate for each of Fanchuan and Kookmin. This is done so that any improvements are due to the per-batch adjustment, rather than due to larger (overall) learning rates (i.e. simply discovering a better hyperparameter setting accidentally). Further, we keep the ``structure'' of Fanchuan and Kookmin intact; i.e. the same number of epochs and structure that dictates how the erase and repair phases are arranged. We only update the objective function of repair. There are of course other forms of stitching that we hope future work explores.

\begin{figure}
    \centering
    \includegraphics[scale=0.2]{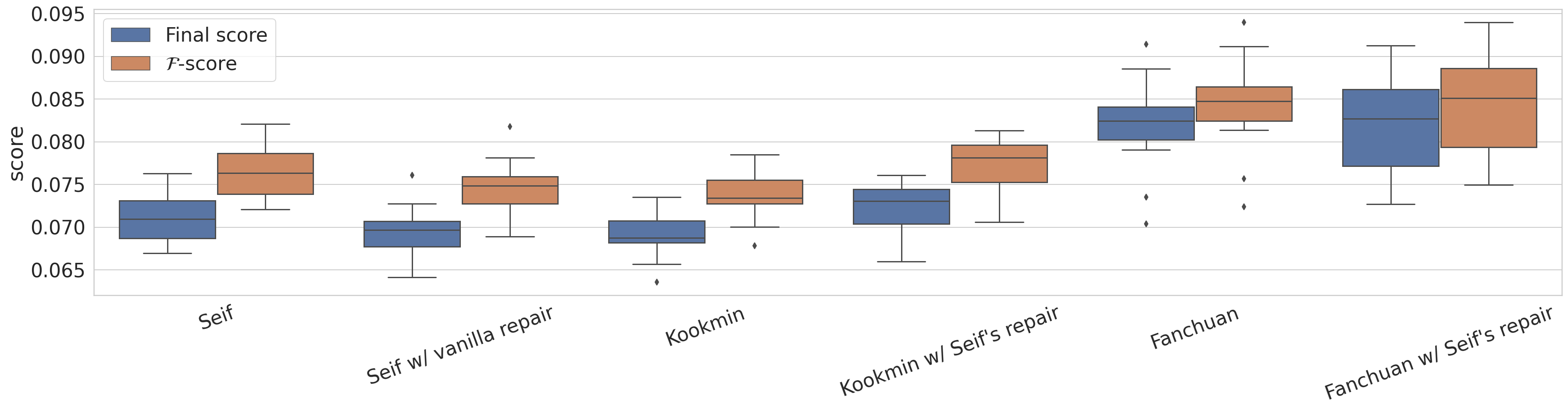}
    \caption{Exploring direct stitching of different erase and repair phases. We used Setup ``Reuse-$N$-$N$'' here, with $N$ = 1024, $E$ = 1024, which we verified is a good proxy for Setup ``Full''.}
    \label{fig:erase_repair_stitch}
\end{figure}

We present the results in Figure \ref{fig:erase_repair_stitch}. We find that, first of all, replacing Seif's repair phase with a vanilla repair phase (and making the appropriate learning rate adjustment to keep the vanilla learning rate the same as the average per-batch learning rate of Seif's original recipe) leads to a small degradation to Seif's performance, both in terms of $\mathcal{F}$-score and final score, though within the confidence intervals. Then, when stitching Seif's repair phase into Kookmin and Fanchuan as described above, we observe positive results in the former case and negative in the latter. Interestingly, Fanchuan's confidence intervals become wider with this change, which warrants further investigation. We hypothesize that the structure of Fanchuan, whose second phase iterates between ``erase'' and ``repair'', rather than performing the two phases sequentially, with more epochs each, is potentially more prone to instability that may be caused by using different effective learning rates for different batches. However, the fact that we see a mild improvement for Kookmin by replacing its repair phase by Seif's is encouraging evidence for continuing to explore this direction.

We remark that this type of stitching is the simplest possible and is meant to serve as a proof of concept for pathways of improving existing algorithms by combining insights across submissions. We expect that re-tuning all relevant hyperparamters (of both newly-combined ``erase'' and ``repair'' phases) after stitching would yield stronger results. We hope future work conducts these experiments, towards discovering even better unlearning algorithms.

\subsection{Relationship between \texorpdfstring{$\mathcal{F}$}{F}-scores and a simple MIA} \label{sec:simple_mia}

In this section, we investigate the relationship between our $\mathcal{F}$-score and a different proxy for forgetting quality: using a simple Membership Inference Attack (MIA), referred to as the ``Baseline MIA'' in \cite{kurmanji2024towards}.

This MIA is a binary classifier trained to distinguish the losses of the unlearned model for forget set examples from the losses of the unlearned model on test set examples. Intuitively, if the forget and test sets follow the same distribution, then ideal unlearning according to this metric is given by a 50\% accuracy of this classifier, signalling its inability to tell apart examples that were unlearned from those that were never trained on in the first place (from their losses), and marking a success for unlearning.

Given that the distributions of the forget and test set are not necessarily the same (and thus, even for a perfectly unlearned model, the loss distributions may differ slightly on those two sets), instead of defining ideal unlearning as 50\% accuracy of that classifier, we define it as matching the reference point of the classifier's accuracy when applied on a model retrained-from-scratch. This leads to the ``MIA gap'' score, defined as the absolute difference in the accuracy of the binary classifier when applied to the unlearned and retrained model. Lower is better and the ideal score is 0.

In practice, we implement the training of the binary attacker via cross-validation: it is trained on a subset of forget and test losses, and its success in telling apart those two sets is then evaluated on held-out forget and test losses. We used 10 such splits of the two sets of losses, and we report the score as the average over the held-out losses of each such split.

\begin{figure}
    \centering
    \includegraphics[scale=0.2]{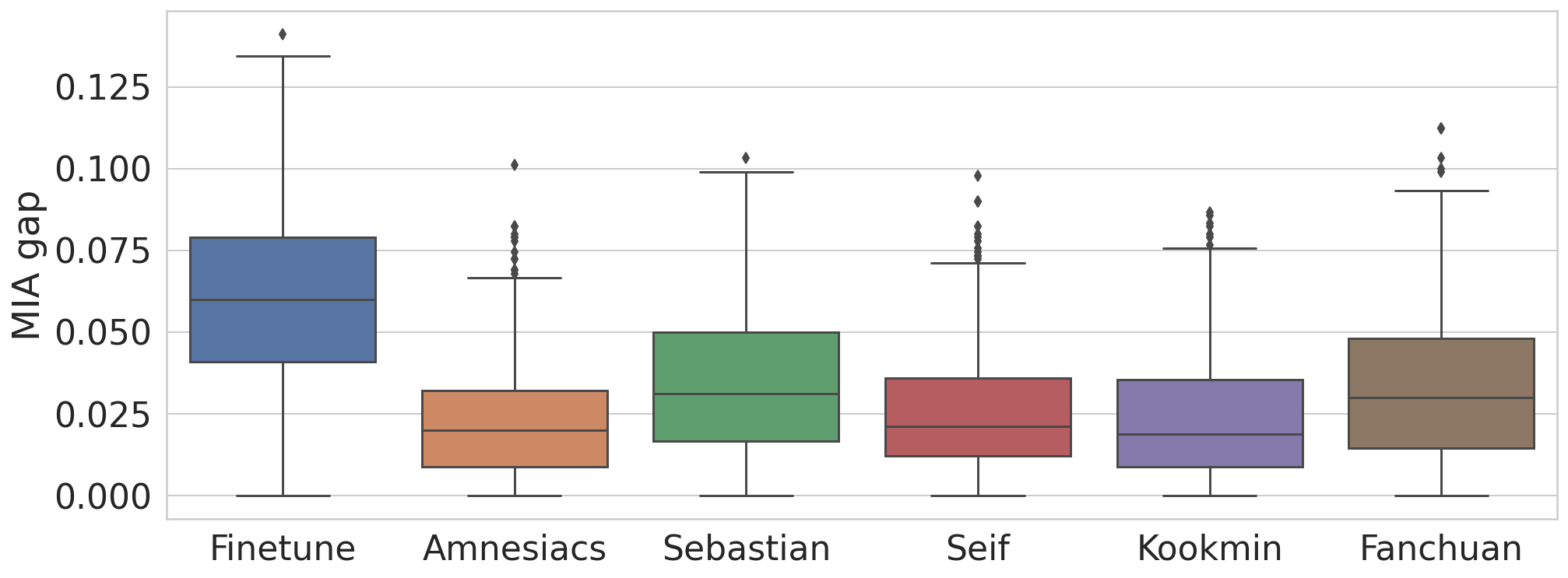}
    \caption{The ``MIA gap'' using a simple membership inference attack (lower is better) for top competition algorithms.}
    \label{fig:simple_mia}
\end{figure}

Further, we repeat the procedure of applying this attack 500 times for retrain-from-scratch (each of those 500 times, the attack is applied on a different sample of retraining-from-scratch) and similarly 500 times for unlearning (each time applied on an unlearned model produced by running the given unlearning procedure on a different sample of the original model). We use these 500 estimates of the MIA score to produce the MIA gap, which we report in Figure \ref{fig:simple_mia}. 

Due to very large confidence intervals, we are not able to make strong claims about the correlation of this metric with our $F$-scores, however Figure Figure \ref{fig:simple_mia} suggests that i) competition methods improve upon the Finetune baseline in this metric too, and that ii) the relative ranking between competition algorithms is not the same as the one produced by our forgetting quality score, according to which Sebastian significantly outperforms other methods on this aspect. We leave it to future work to investigate whether these observations hold and are statistically significant when the variance of MIA gap scores is reduced.

\subsection{Histograms of unlearned and retrained distributions} \label{sec:histogram_visualizations}
\begin{figure}[ht]
\centering
\includegraphics[scale=0.27]{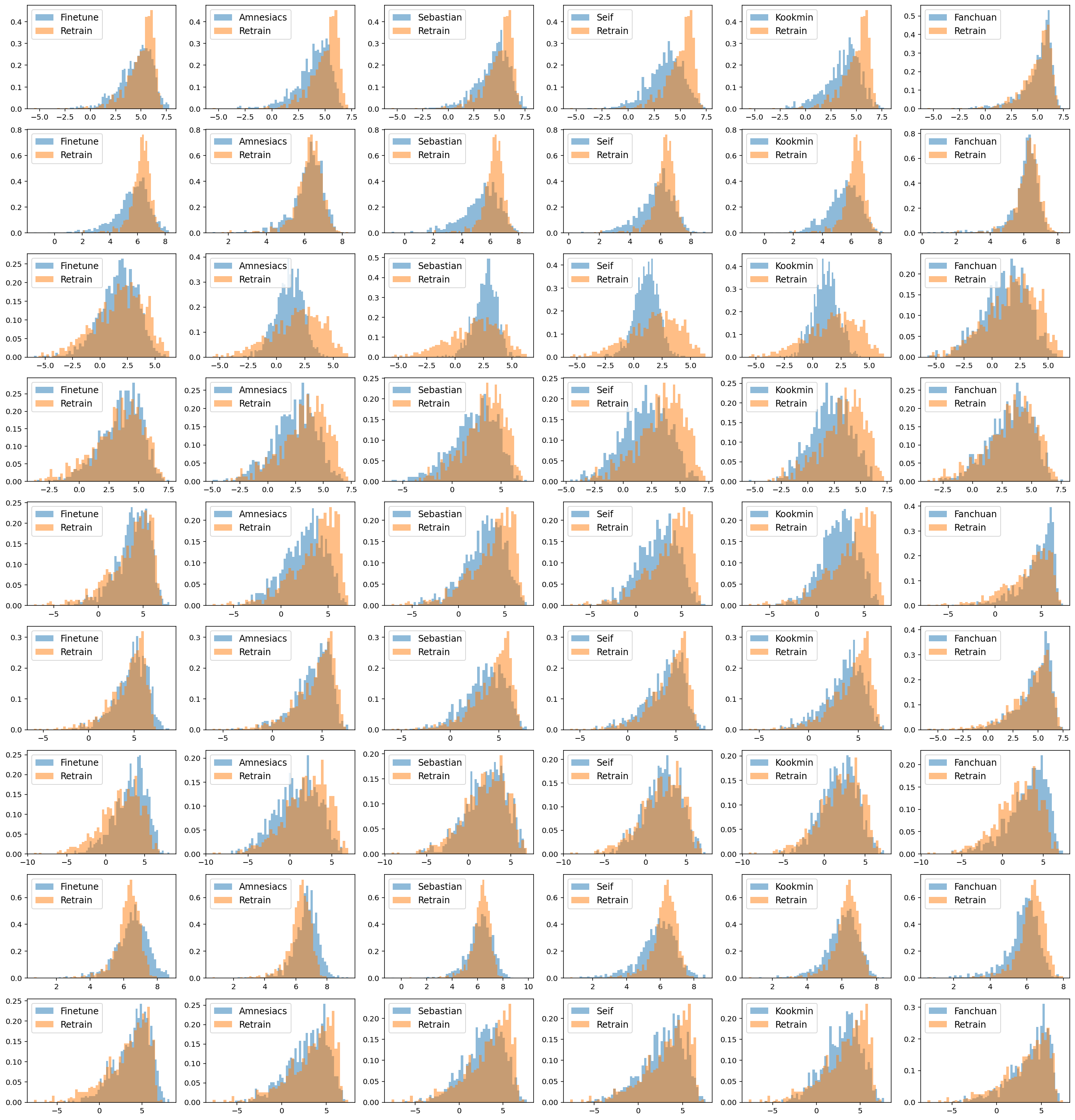}
\caption{Histograms of our one-dimensional test statistic (the logit-scaled probability of the correct class) for different forget set examples and the five top unlearning algorithms, as well as the Finetune baseline. Each row represents a different forget set example and each column a different unlearning algorithm.}
\label{fig:unlearned_retrained_hist}
\end{figure}

In Figure \ref{fig:unlearned_retrained_hist}, we plot the histograms of our one-dimensional test statistic (the logit-scaled probability of the correct class) for different forget set examples and the five top unlearning algorithms, as well as the Finetune baseline. Each row represents a different forget set example and each column a different unlearning algorithm. We observe that, while different unlearning algorithms lead to different distributions, we sometimes observe per-example patterns across algorithms, especially across Kookmin, Seif, Sebastian and Amnesaics. This is particularly visible in rows 3-5, for instance. Finally, these plots illustrate the diversity of shapes that these histograms can take, making it difficult to separate unlearned from retrained distributions well with a single-threshold decision rule, in all cases. Note that, in the case where one distribution is more peaky that the other (e.g. Sebastian, in the second-to-last row), a double-threshold rule, that attempts to enclose the peakiest of the two distributions is better than any single-threshold one.

\subsection{Exploring the degree of overfitting the attacker} \label{sec:degree_of_overfitting} As we described in Section \ref{sec:forget_quality}, we compute a $\varepsilon$ for each example $s \in \forgetset$ by running $m$ attacks to separate the (post-processed through $h$) distributions of outputs of the retrained and unlearned models, when given as input example $s$. The procedure we outlined for that (see Section \ref{sec:forget_quality} and Algorithm \ref{alg:eps-from-fpr-fnr}) determines the instantiation of the attack to use (i.e. the specific choice for the threshold $t$ or $t_1$ and $t_2$) based on how well it separates the $N$ samples from each of the two distributions. Then, the reported $\varepsilon$ for that example is based on the success of that instantiated attack in telling apart those same $N$ samples from each distribution.

An alternative framework disentangles the following two steps: i) the step of ``fitting'' the attacker, i.e. instantiating a decision rule through choosing a threshold or pair of thresholds, and ii) the step of  ``evaluating'' the attacker, i.e. computing $\varepsilon$ based on the success of the chosen decision rule. Importantly, a different set of $N$ samples from each distribution would be used for each of those two steps. This disentanglement intuitively will prevent the attacker from ``overfitting'' to the $N$ samples from the two distributions: that is, picking a decision rule that happens to separate well the specific $N$ samples from the two distributions, but would not work as well if the two distributions were estimated with a different set of $N$ samples each.

In this section, we examine the degree of potential overfitting of the attacker by disentangling the two steps as mentioned above, and using separate sets of $N$ samples for each. This, however, comes with the following technical difficulty. Notice from Algorithm \ref{alg:eps-from-fpr-fnr} that we discard a decision rule if exactly one of its FPR or FNR is 0. This is based on a hypothesis that the FPR or FNR is 0 only due to the fact that have a limited number of samples of a continuous distribution and that, having had more samples from it, the FPR / FNR may have been small but not exactly 0. Now, when disentangling the two steps mentioned above, step i may yield a decision rule that while is considered valid for the ``fitting'' set of samples, is considered invalid (one of FNR or FPR is 0) when applied on the ``evaluation'' set of samples, leaving us unable to compute a $\varepsilon$ based on the chosen decision rule.

To alleviate this issue, we apply Kernel Density Estimation (KDE) to smooth the distribution in step 2. This is useful as the estimated PDF will have some (small) probability mass even in areas where the raw empirical distribution does not. Therefore, this makes it unlikely that a threshold chosen based on the ``fitting'' samples will yield an FPR of FNR of exactly 0 on the smoothed version of the evaluation distribution. 

\begin{figure}
    \centering
    \includegraphics[scale=0.25]{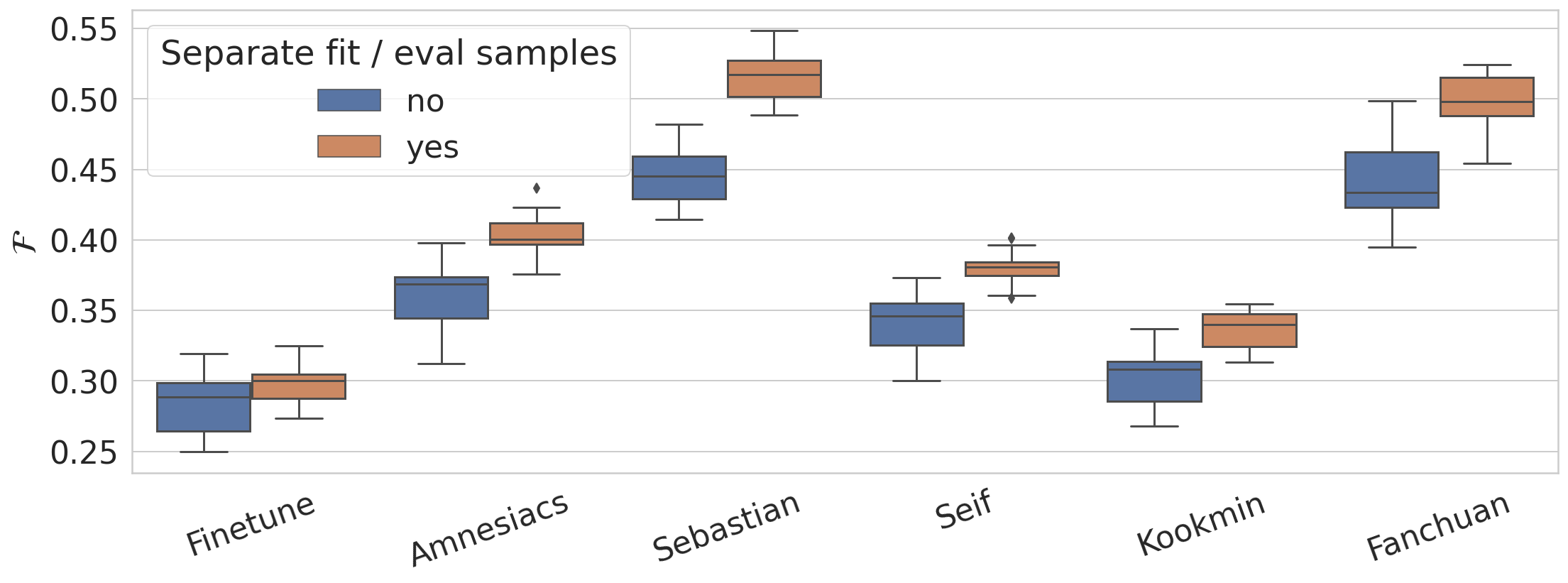}
    \caption{The degree of attacker ``overfitting''. We compare our default setup, as described in \ref{sec:forget_quality} and Algorithm \ref{alg:eps-from-fpr-fnr} against a setup that disentangles (and uses separate sets of $N$ samples from the retrained and unlearned distributions for) two steps: i) fitting the attacker, i.e. instantiating a decision rule by picking a threshold or pair of thresholds, and ii) evaluating the attacker, i.e using that fixed decision rule to compute FPR and FNR and, from those ingredients, obtain $\varepsilon$. As discussed in Section \ref{sec:degree_of_overfitting}, when disentangling the two steps, we use KDE to smooth the evaluation distribution, in step ii. Intuitively, allowing the attacker to overfit on the set of samples used for ``fitting'' will yield results that are more optimistic for the attacker (and thus less optimistic for unlearning, producing smaller $\mathcal{F}$ values). Indeed, we observe some degree of overfitting: the $\mathcal{F}$ values are larger when using a fresh set of samples from each distribution to evaluate the chosen decision rule. However, we find that this effect isn't severe and does not change the relative rankings between different unlearning algorithms.}
    \label{fig:degree_of_overfitting}
\end{figure}

We present the results of this investigation in Figure \ref{fig:degree_of_overfitting}, using setup ``Full'' (see Figure \ref{fig:setups}) and $N = 1024$. When the two phases are disentangled, each of the two phases receives 1024 samples from each distribution. As expected, we do observe some degree of overfitting: the $\mathcal{F}$ values are larger when using a fresh set of samples from each distribution to evaluate the chosen decision rule. However, we find that this effect isn't severe and does not change the relative rankings between different unlearning algorithms.

\subsection{Exploring the effect of \texorpdfstring{$N$}{N}} \label{sec:effect_of_N} We now discuss how the results change as we modify $N$, the number of samples that are used to instantiate the empirical distributions of unlearning and retraining on which we base our evaluation.

We decided to conduct this investigation in the setup of disentangled sets of samples for ``fitting'' and ``evaluating'', as is described in Section \ref{sec:degree_of_overfitting}. This is because, varying $N$ with fitting and evaluation entangled makes it harder to interpret the results: increasing values of $N$ in that setting would lead to two effects simultaneously: i) a stronger attacker that has more samples to inform its decision about the ideal threshold(s), and ii) more accurate evaluation due to a better estimation of the two distributions.

\begin{figure}
    \centering
    \includegraphics[scale=0.25]{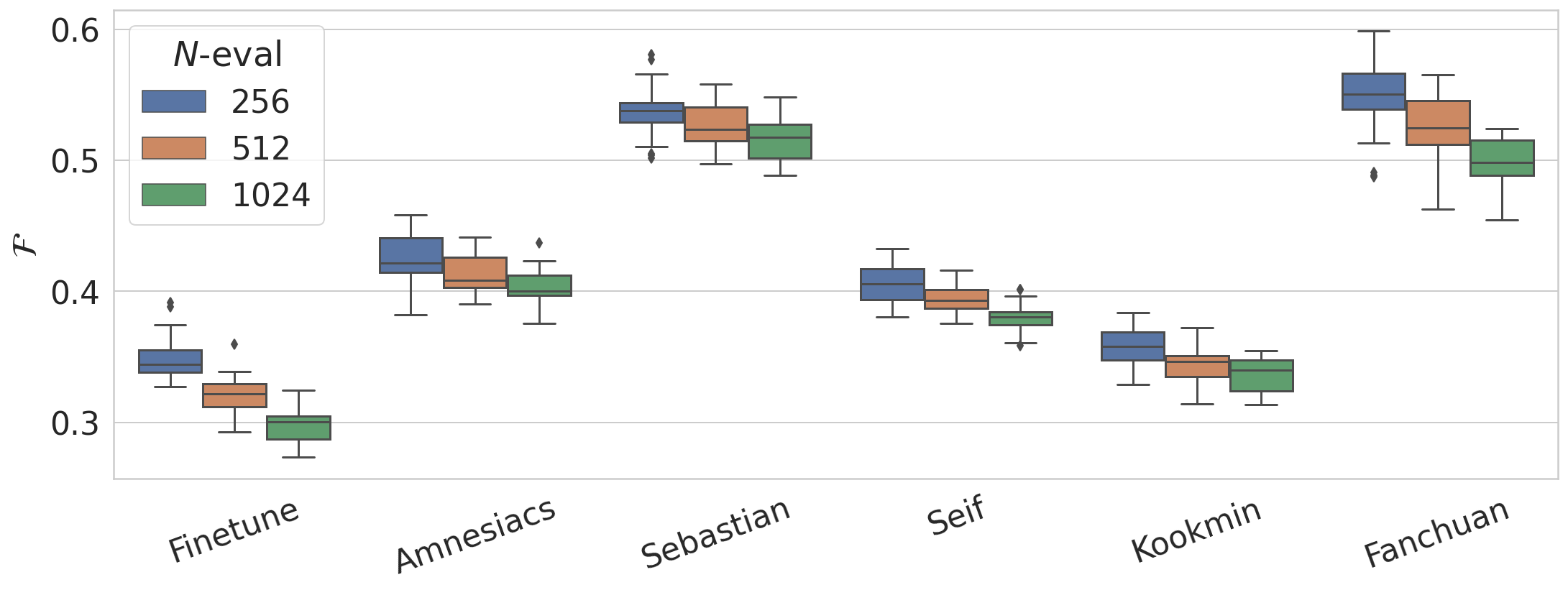}
    \caption{Investigating the effect of the number of samples from each distribution used for evaluation, $N$-eval, in terms of the forgetting quality estimate $\mathcal{F}$. In all cases the fitting step was performed using 1024 samples from each of the retrained and unlearned distributions (see Section \ref{sec:degree_of_overfitting} for a description of the two-step setup of disentangled ``fitting'' and ``evaluation'').}
    \label{fig:effect_of_N}
\end{figure}

In Figure \ref{fig:effect_of_N}, we compare the $\mathcal{F}$ scores obtained using 1024 samples from each distribution for fitting, and $N$-eval samples for evaluation, for different values of $N$-eval.  We observe that, while the differences are not severe and don't affect the relative ranking between different algorithms, larger values of $N$-eval, associated with a better representation of the two distributions, do lead to slightly lower means of $\mathcal{F}$ values, though well within the confidence intervals of other $N$-eval values.

\subsection{Exploring bootstrapping to efficiently estimate \texorpdfstring{$\mathcal{F}$}{F}} \label{sec:bootstrapping} 
As discussed in Section \ref{sec:practical_instantiations}, we explored using bootstrapping, a common statistical tool, to reduce the computation cost of Setup ``Full'' by sampling with replacement from a smaller pool of samples than those required to compute the ``Full'' variant. 

Specifically, from a pool size of $K$ triplets of $(\theta^o, \theta^u, \theta^r)$, we sample $N$ of them with replacement, and compute an estimate of $\mathcal{F}$. We repeat this procedure $E$ times, yielding a total of $E$ estimates of $\mathcal{F}$, as in Setup ``Full'', but requiring only $K$ models from each distribution here for all $E$ estimates, rather than $N \times E$ as in Setup ``Full''. In both cases, we set $E$ to 20 in our experiments, unless otherwise specified. 

\begin{figure}
    \centering
    \begin{subfigure}[b]{0.9\textwidth}
         \includegraphics[scale=0.22]{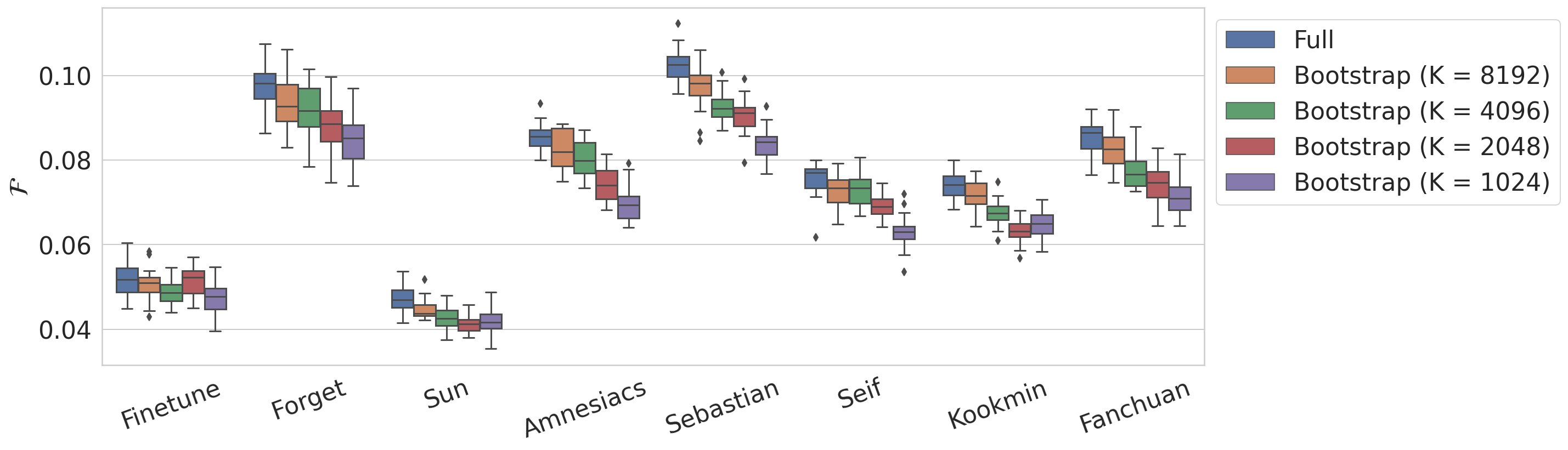}
         \caption{Investigation of different values of $K$ (for fixed $E = 20$). As expected, we observe that, the larger $K$ is, the better bootstrapping can estimate the Full setup, since more samples from the population are used, rather than simply reusing repetitions of a smaller sample size.}
         \label{fig:boot_K}
    \end{subfigure}
    \begin{subfigure}[b]{0.9\textwidth}
         \includegraphics[scale=0.22]{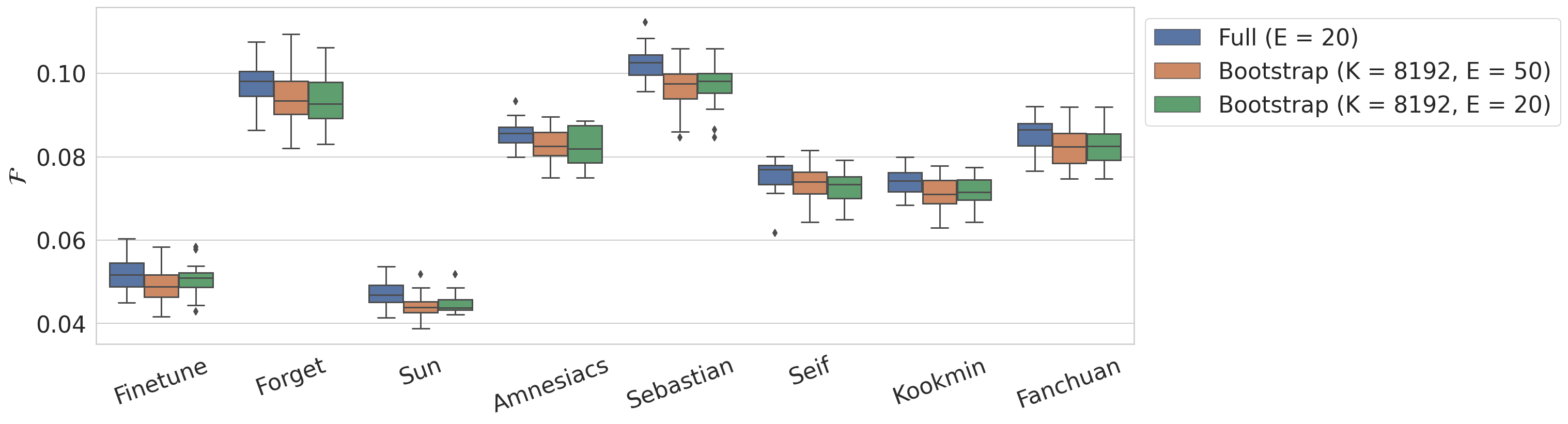}
         \caption{Investigation of different values of $E$ (for fixed $K = 8192$). We observe that increasing $E$  (i.e. simply sampling with replacement more times) does not further improve the approximation. }
         \label{fig:boot_E}
    \end{subfigure}
    \caption{Can we use bootstrapping to estimate $\mathcal{F}$ more efficiently?}
    \label{fig:more_results}
\end{figure}

We report results for different values of the pool size $K$ in Figure \ref{fig:boot_K} and, for the largest pool size, we report results for different values of $E$ in Figure \ref{fig:boot_E}. As expected, we observe that, the larger $K$ is, the better bootstrapping can estimate the ``Full'' setup, since more samples from the population are used, rather than simply reusing repetitions of a smaller sample size. We note that, perhaps surprisingly, we can get a decent approximation of setup ``Full'' via bootstrapping with $K$ = 8192, which yields significant savings in the compute cost, especially for larger values of $E$: it requires training a total of $N \times 8$ models per distribution in total, whereas setup ``Full'' requires $N \times E$, saving more than half of the compute cost for our default value $E = 20$. We also observe that, for the largest pool size, increasing $E$ for bootstrapping (i.e. simply sampling with replacement more times) does not further improve the approximation.

\begin{figure}
    \centering
     \includegraphics[scale=0.22]{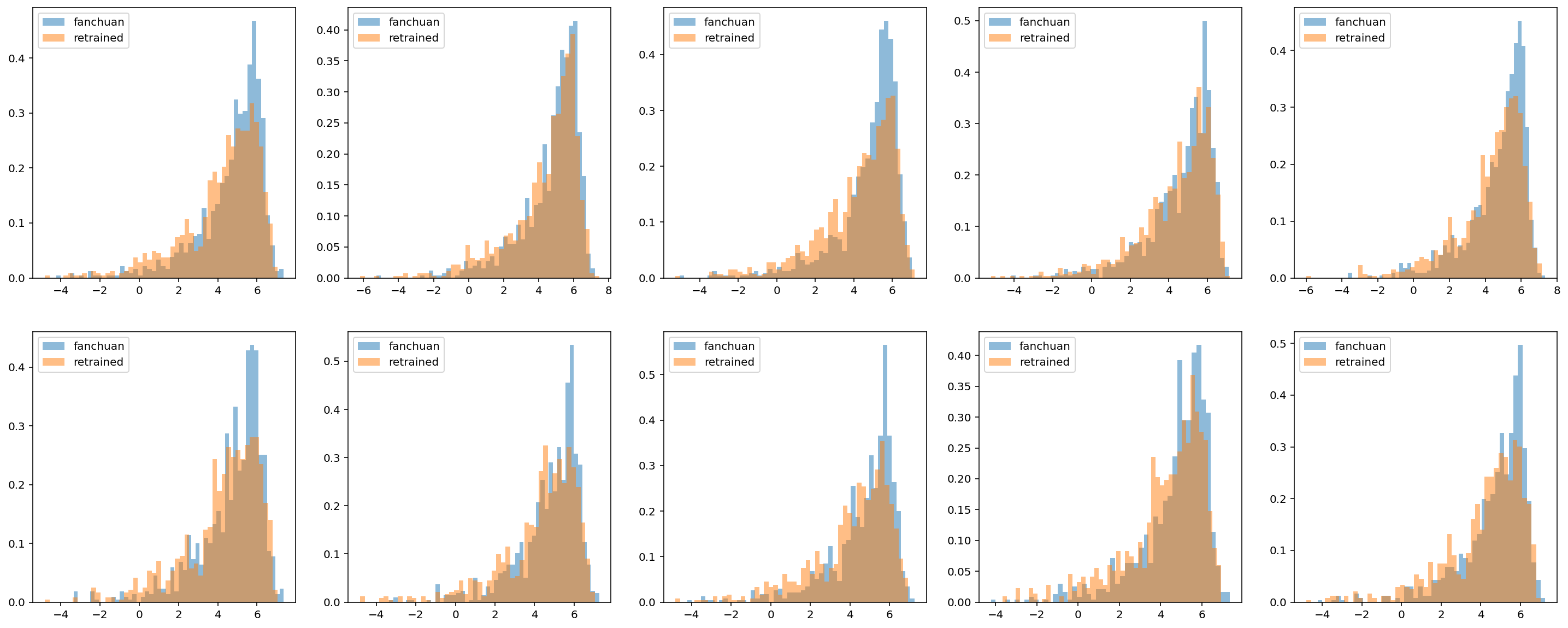}
     \caption{Qualitative investigation of the unlearned and retrained distributions (of our test statistic, i.e. logit-scaled confidence) obtained from fresh versus bootstrapped samples, for a particular forget set example, and for the unlearning algorithm Fanchuan. Specifically, each subplot in the top row shows the empirical distributions instantiated through a different set of $N = 1024$ fresh samples from the two distributions (unlearned versus retrained). In contrast, all subplots in the bottom row are the empirical distributions of different bootstrapped samples, all originating from the pool corresponding to the samples of leftmost subplot of the top row (so $K = 1024$). We observe that, even with this smallest considered pool size, the bottom row does not look too different visually from the top row, offering some additional qualitative evidence for the ability of bootstrapping to mimic the unlearned and retrained distributions of interest.}
     \label{fig:boot_qualitative}
\end{figure}

We also show a visual illustration of the distributions (of our test statistic, i.e. logit-scaled confidence) obtained from $N = 1024$ fresh samples (top) versus $N = 1024$ bootstrapped samples (from the smallest pool size considered, $K = 1024$) for unlearning method Fanchuan and for a particular example in the forget set, in Figure \ref{fig:boot_qualitative}. We notice that, even when considering $K = 1024$, the bottom row (bootstrapped samples) don't appear too different from the histograms in the top row, offering some additional qualitative evidence for the ability of bootstrapping to mimic the unlearned and retrained distributions of interest.

\end{document}